\RequirePackage{fix-cm}
\documentclass[smallextended]{svjour3}       
\smartqed  
\usepackage{comment}
\usepackage{times}
\usepackage{epsfig}
\usepackage[numbers,sort&compress]{natbib}
\usepackage[caption=false]{subfig}
\usepackage{lipsum}
\usepackage{graphicx}
\usepackage{amsmath}
\usepackage{amssymb}
\usepackage{caption}
\usepackage{xcolor}
\usepackage{tabularx}
\usepackage{soul}
\usepackage{multirow}

\begin{document}

\title{On the Robustness of Human Pose Estimation
}
\subtitle{}


\author{
$^1$Sahil Shah\thanks{$^{\dagger}$ Equal Contribution}$^{\dagger}$ $^1$Naman Jain$^{\dagger}$ $^2$Abhishek~Sharma $^3$Arjun~Jain\vspace{.2cm}\\
}


\institute{$^1$ 
              Department of Computer Science, IIT Bombay \\
              \email{\{sahilshah, namanjain\}@cse.iitb.ac.in}  \\    $^2$ Axogyan AI \\
              \email{abhisharayiya@gmail.com \\
              $^3$ Indian Institute of Science, Axogyan AI \\
              \email{arjunjain@iisc.ac.in}}
}

\date{}

\authorrunning{Sahil Shah$^{\dagger}$, Naman Jain$^{\dagger}$, Abhishek Sharma, Arjun Jain}

\maketitle

\begin{abstract}
This paper provides a comprehensive and exhaustive study of adversarial attacks on human pose estimation models and the evaluation of their robustness.  Besides highlighting the important differences between well-studied classification and human pose-estimation systems w.r.t. adversarial attacks, we also provide deep insights into the design choices of pose-estimation systems to shape future work. We benchmark the robustness of several 2D single person pose-estimation architectures trained on multiple datasets, MPII and COCO. In doing so, we also explore the problem of attacking non-classification networks including regression based networks, which has been virtually unexplored in the past. 
\par We find that compared to classification and semantic segmentation, human pose estimation architectures are relatively robust to adversarial attacks with the single-step attacks being surprisingly ineffective. Our study shows that the heatmap-based pose-estimation models are notably robust than their direct regression-based systems and that the systems which explicitly model anthropomorphic semantics of human body fare better than their other counterparts. Besides, targeted attacks are more difficult to obtain than un-targeted ones and some body-joints are easier to fool than the others. We present visualizations of universal perturbations to facilitate unprecedented insights into their workings on pose-estimation. Additionally, we show them to generalize well across different networks. Finally we perform a user study about perceptibility of these examples.
\end{abstract}

\section{Introduction}
We are witnessing an exponential growth in the deployment of deep-learning based systems for real-world automation. The applications include, but not limited to, autonomous driving, medical-image analysis, security and surveillance, visual-signal driven human-computer-interaction such as eye-tracking, human-pose estimation and egocentric data-analysis. Such systems require to serve the entire spectrum of variations in human population and operating conditions, while maintaining a high-level of accuracy. Hence, not only the robustness against naturally occurring variations but also against adversarial attacks are critically required for real-world deployment. Unfortunately, deep-learning systems have been proven to be extremely prone to adversarial attacks in the form of imperceptible noise added to the input\cite{AdvGAN, XieAdversarialDetection, JianjunAdversarialDetection, KurakinADVERSARIALSCALE, Kurakin2017ADVERSARIALWORLD}. Given that adversarial attacks can break such systems, the robustness against adversarial attacks \emph{must} be considered as critical a metric as accuracy, computation, generalization and/or interpretation. 

Ever since the discovery of adversarial attack, studying their effects has received significant attention. While some systems, such as classification ~\cite{CarliniTowardsNetworks, PapernotTheSettings, LiuDELVINGATTACKS, KurakinADVERSARIALSCALE, BastaniMeasuringConstraints, DongBoostingMomentum, PoursaeedGenerativePerturbations, Arnab_2018_CVPR, Fischer2017ADVERSARIALSEGMENTATION}, have witnessed more attention over the others, such as regression ~\cite{SongDetec,FasterRCNNattack,ranjan2019attacking}, it's important to underscore that the characteristics of adversarial attacks don't generalize across different applications. It's due to the fact that different applications require carefully designed deep-learning systems with unique components and embedded domain-knowledge. Therefore, a careful \emph{application-specific} study of adversarial attacks is required as a first step towards building robustness.

Recently, Human-pose estimation, referred to as \textbf{HPE} for brevity, has emerged as an important HCI component with an aim to deploy HPE on commodity hardware. HPE is an interesting deep-learning system that uses a blend of regression and classification approaches along with anthropometric knowledge to estimate human-body pose. Since humans are the central object of attention for any HPE systems, it makes HPE systems extremely susceptible to adversarial attacks significantly different from an object-detection and/or semantic-segmentation. To this end, we employ the contemporary state-of-the-art adversarial attack algorithms and design multiple HPE-specific attacks on several state-of-the art HPE systems to facilitate the first comprehensive study of the effects of adversarial attacks on HPE. Our analysis on two large-scale benchmark datasets, MPII~\cite{andriluka14cvpr} and COCO~\cite{coco}, reveals interesting insights about the effect of adversarial attacks w.r.t. different system-design choices, like heatmaps vs. direct regression, multi-scale processing, attention and anthropometric constraints.

\begin{figure*}[tbh]

\centering
  \subfloat[Target Pose \label{fig:target}]
  {
  \includegraphics[width=0.13\linewidth,height=0.254\linewidth]{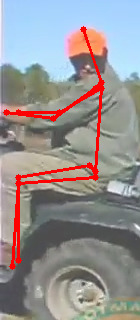}
  }\subfloat[ Attention-HG \label{fig:att}]
  {
  \includegraphics[width=0.13\linewidth]{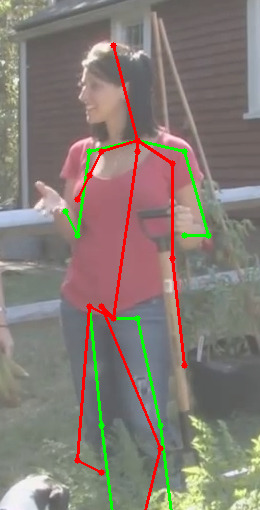}
  }
  \subfloat[8-Stacked-HG\label{fig:8shg}]
  {
  \includegraphics[width=0.13\linewidth] {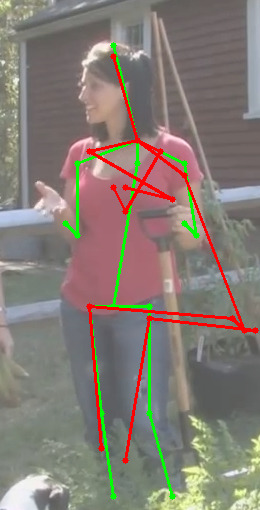}
  }
  \subfloat[DeepPose\label{fig:deeppose}]
  { 
  \includegraphics[width=0.13\linewidth]{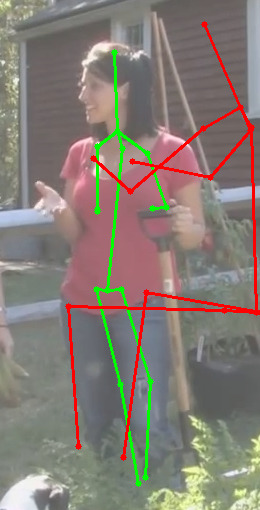}
  }
    \subfloat[Chained-Preds\label{fig:chain}]
  {
  \includegraphics[width=0.13\linewidth]{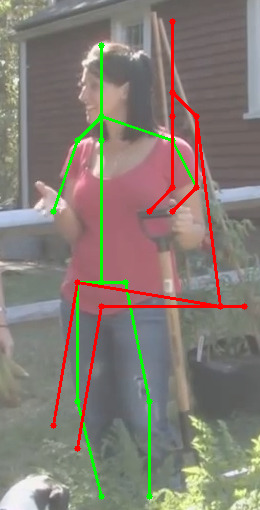}
  }
  \subfloat[DLCM\label{fig:dlcm}]
  {
  \includegraphics[width=0.13\linewidth]{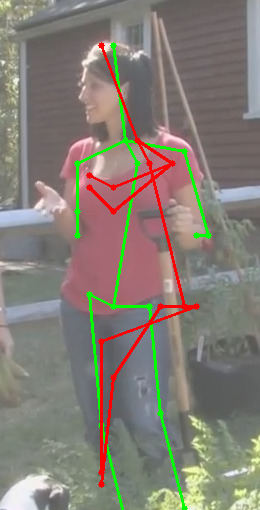}
  }
  \subfloat[2-Stacked-HG\label{fig:dlcm}]
  {
  \includegraphics[width=0.13\linewidth]{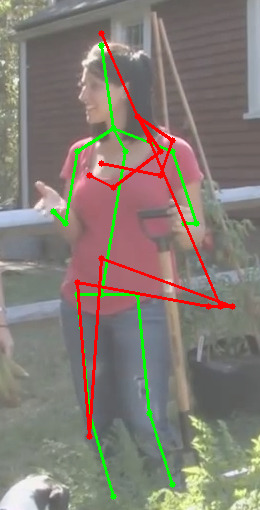}
  }
  \caption{Example of various targeted adversarial attacks of different networks on the MPII benchmark. (a) represents the target pose used for computing the adversarial perturbation while in figures (b-g) : Green skeletons show the original predictions while the red skeletons show the predictions for perturbed image.}
  \label{fig1}  
  \vspace{-11pt}
\end{figure*}

Some of the obtained insights are similar in nature w.r.t. the studies of adversarial attacks on image-classification ~\cite{CarliniTowardsNetworks, PapernotTheSettings, KurakinADVERSARIALSCALE}, object-detection ~\cite{SongDetec, FasterRCNNattack} and semantic-segmentation ~\cite{Arnab_2018_CVPR, Fischer2017ADVERSARIALSEGMENTATION} systems. For example- ImageNet pre-training improves robustness ~\cite{pretrain, Chen_2020_pretrain_CVPR}, multi-scale processing improves robustness ~\cite{Arnab_2018_CVPR, ranjan2019attacking}. While the aforementioned insights generalize across different applications, below we enlist some HPE-specific insights obtained from the study presented in this article-
\begin{itemize}
    \item Heatmap-based HPE systems ~\cite{TompsonJointEstimation, NewellStackedEstimation, TangDeeplyEstimation} are significantly more robust than direct regression-based systems ~\cite{ToshevDeepPose:Networks}.
    \item Employing anthropometric compostion/constraint ~\cite{TangDeeplyEstimation} improves the robustness against adversarial attacks.
    \item We propose HPE-specifc \emph{targeted} and \emph{un-trageted} attacks and show that the former are harder to execute and require careful tuning of hyper-parameters.
    \item We show that the attacks injected deeper into the network are more detrimental than the ones applied at the last layer only, which is intuitive.
    \item Universal perturbations ~\cite{Moosavi_DezfooliUniversalPerturbations, Metzen_2017_ICCV} are extremely effective against HPE systems. Their visualization reveals that they hallucinate body-joints, like limbs, head and shoulder, all across the input image which effectively confuses the network, and therefore, this attack generalizes fairly across different networks and system-designs. Moreover, the skeletons predicted under universal attacks tend to resemble the same pose even in the absfence of any explicit constraint to do so.
    \item Among different body-joints, the hip and the joints below the hip are most vulnerable, while head the neck are the most robust against adversarial attacks.
    \item We also evaluate the robustness of 2D multi-person and single person 3D pose estimation networks.  
    \item We compare the performance of bottom-up and top-down approaches for 2D multi-person HPE and find that under extreme attacks bottom-up methods predict a surprisingly large number of humans.  
    \item Finally, we initiate a user study to analyse the perceptibility of our attacks and show that these attacks do not interfere with human performance.
\end{itemize}
We hope that the insights presented in this article will motivate future research and pave the way for the development of robust real-world HPE systems. This work is an extension of our previous work workshop submission \cite{our_workshop}, which is made more presentable and self-contained, supplanted with - additional studies into 3D and multi-person pose estimation systems, discussions around the quality of heatmaps, user-study to analyze the visual imperceptibility of the adversarial attacks, and additional visualizations.

\section{Related Work}
    
Immediately after the inception of deep-learning, in the form of AlexNet~\cite{imagenet}, \cite{SzegedyIntriguingNetworks} showed that deep-neural nets are easily fooled by noise generated using second-order optimization, L-BFGS in this case, via back-propagation. Later,~\cite{GoodfellowEXPLAININGEXAMPLES} introduced Fast-Gradient-Sign-Method, or FGSM for brevity, which is a first-order gradient method and hence more efficient than L-BFGS based adversarial noise generation. FGSM was further developed into Iterative-Gradient-Sign-Method (IGSM)~\cite{Moosavi-DezfooliDeepFool:Networks} that takes multiple FGSM steps, which was later developed to optimize for the least likely class in ~\cite{Kurakin2017ADVERSARIALWORLD}. Since then this field has witnessed active research that has led to the extension of adversarial attacks with different datasets, penalty functions and optimization methods~\cite{CarliniTowardsNetworks, KurakinADVERSARIALSCALE, DongBoostingMomentum, PapernotTheSettings,  LiuDELVINGATTACKS, Moosavi-DezfooliDeepFool:Networks, BastaniMeasuringConstraints, CisseHoudini:Examples, NguyenDeepImages, svmattacks, Su_2018_ECCV}. An altogether different line of work employed DNNs to directly generate adversarial perturbations from an input image \cite{DBLP:BalujaATN, SarkarUPSETClassifiers, PoursaeedGenerativePerturbations, AdvGAN}. These approaches, however, require complete access to the inference network that limits their practicality for real-world application. Black-box attacks ~\cite{LiuDELVINGATTACKS, paperblack, PapernotTheSettings}, on the other hand, generalize across networks and do not need access to the target network that makes them more practical. 

Most of the aforementioned attacks are image-specific and need costly back-propagation through the entire network. To mitigate this issue, universal adversarial perturbations~\cite{Moosavi_DezfooliUniversalPerturbations, Metzen_2017_ICCV} were proposed that can be learned for a particular network and can be applied to any image to fool the network. The effectiveness of the universal adversarial perturbations was shown on ImageNet in~\cite{Moosavi_DezfooliUniversalPerturbations}, while~\cite{Metzen_2017_ICCV} analyzed the same for semantic segmentation. In the past, the study of adversarial attacks has mostly been limited to image classification. Recently, however, such attacks have been analyzed for other practical problems as well, such as image segmentation (again a per-pixel \emph{classification})~\cite{PoursaeedGenerativePerturbations, Xiao2018CharacterizingSegmentation, Metzen_2017_ICCV, Arnab_2018_CVPR, Fischer2017ADVERSARIALSEGMENTATION, XieAdversarialDetection}, object detection~\cite{SongDetec, FasterRCNNattack}, visual question answering~\cite{XuFoolingMechanism} and/or optical flow \cite{ranjan2019attacking}.

Human pose estimation (HPE), unfortunately, hasn't witnessed any systematic effort to analyze the effect of adversarial attacks and the closest work to ours is~\cite{CisseHoudini:Examples} that explores metric specific loss functions for different tasks. It focuses on exploiting a loss function frameworks to develop metric specific attacks and demonstrates the approach for classification, segmentation and HPE. Therefore, this study lacks in-depth analysis of adversarial attacks on HPE systems. We, on the other hand, present a comprehensive analysis of the effects of adversarial attack on the HPE systems to obtain deeper insights that can be useful to create adversarial-robust HPE systems in the future. 

\section{Background, Notations and Experimental Settings} 
\label{sec:background}
This section contains a brief background on deep-learning based single-person 2D-HPE systems and presents an ontology of these methods w.r.t. the loss function and/or use of anthropometric information/constraints along with the details of the HPE systems that we analyze. Next, we present a brief overview of multi-person 2D-HPE systems and single-person 3D-HPE systems to complete the spectrum of body-joint location predicting HPE systems. We further present a brief introduction to adversarial attacks to facilitate better appreciation and understanding of the material presented in this article. We also provide information about the hyper-parameter settings of different adversarial attacks.

\subsection{Single-Person 2D Human-Pose Estimation (HPE) Systems}\label{subsec:HPE}
\begin{figure}[!t]
\centering
\subfloat[DeepPose - A direct regression model ]{\includegraphics[width=0.5\linewidth]{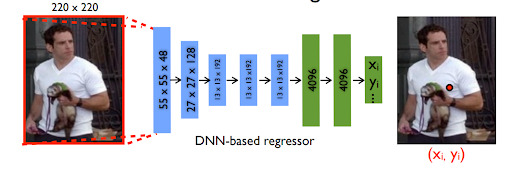}}
\centering
\subfloat[DLCM - A heatmap regression model]{\includegraphics[width=0.5\linewidth]{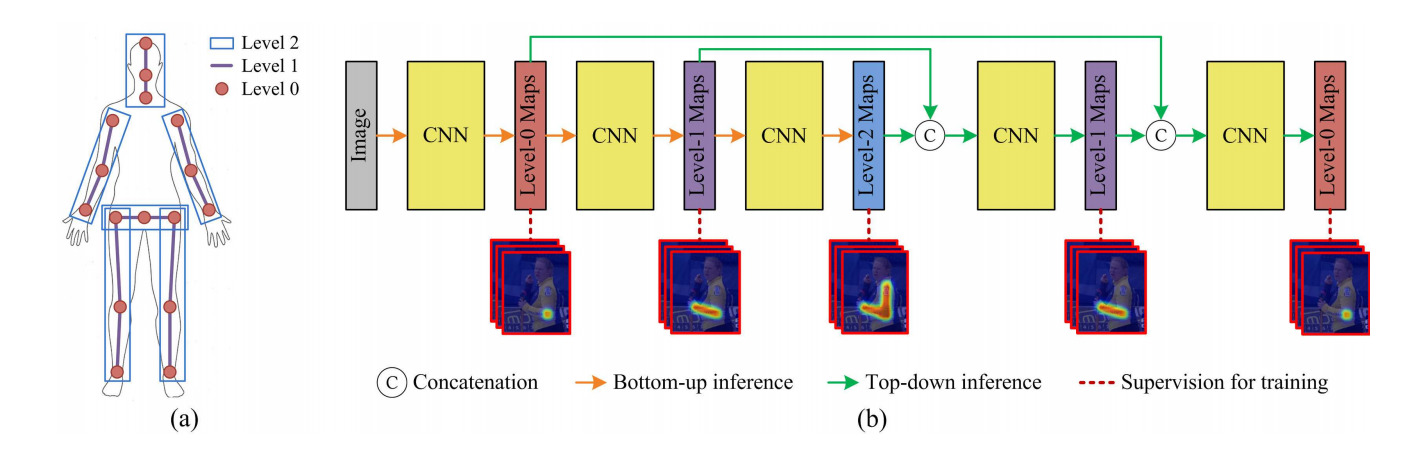}}

\caption{Overview of direct regression and heatmap regression based approaches.}
\label{fig:directvsheatmap}
\end{figure}

2D HPE systems have witnessed a quantum jump in their performance with the use of  ConvNets. One of the first approaches to employ ConvNet features was \emph{DeepPose}~\cite{ToshevDeepPose:Networks} that used a pre-trained AlexNet~\cite{alexnet} to obtain a 4096-dimensional feature vector from an image, $I$, followed by an MLP to regress for the $(x,y)$ coordinates of $k$ body-joints. Later, \cite{JainICLR} introduced a heatmap-based approach for pose-estimation where for training, a multi-scale image patch was used as input and was classified as either a joint such as wrist, elbow, etc. or background. Once trained, this model was then run in a sliding window fashion over the test image to yield heat-maps corresponding to each of joints. This approach was further refined in~\cite{TompsonJointEstimation} by representing the $k$ joint-locations as $k$ output channels, one for each joint, with a Gaussian bump centered at the corresponding joint locations and trained using the entire image as the input. This heatmap-based approach affords effective training with the negative patches (not containing a body joint) that significantly improved the performance over \cite{JainICLR}. Practically, the input image, $I$, is passed through a series of convolution-layers and feature maps at different resolutions are concatenated to finally regress for the ground-truth heatmaps. The approach that directly regresses for the $(x,y)$ coordinates~\cite{ToshevDeepPose:Networks} is knows as \emph{direct regression} while the latter presented in ~\cite{JainICLR, TompsonJointEstimation} is known as \emph{heatmap-based} approach for HPE. These approaches are the standard models for HPE systems. These two approaches are visually explained in Figure~\ref{fig:directvsheatmap}.
Over the years both direct-regression and heatmap-based approaches have witnessed several improvements, most notable ones are employment of Stacked-Hourglasses (or SHG for brevity) \cite{NewellStackedEstimation} as backbone for heatmap HPE and iterative prediction of joints in direct-regression framework (Iterative Error Feedback Method \cite{ief}).  SHGs use a recurring structure of encoder-decoder pair to feed the previously predicted heatmaps for further processing concatenated with the original image features. Owing to it's superb performance, SHGs are the de-facto backbone architecture for HPE systems.

Human-body is not an arbitrary-shaped deforming object, rather, it has an intrinsic anthropometric structure that manifests in the form of bone-length ratios, left-right symmetry, hierarchical structure, joint-angle constraints and pose-priors. The aforementioned approaches for HPE system did not explicitly model any of these constraints. A significant improvement to HPE systems came as the incorporation of aforementioned anthropometric structures in the form of learning signal and/or constraints during training and inference. The incorporation of bone-length constraint to direct-regression approach was employed in \cite{sun2017compositional}. \emph{Chained-Prediction}~\cite{GkioxariChainedNetworks} casts the HPE problem as a sequential joint prediction problem with a series of encoder-decoder networks to predict the joint heatmaps, thus conditioning the prediction of joints w.r.t. the pre-computed joints. Motivated by the success of attention-networks, \emph{Hourglass Attention}~\cite{ChuMulti-ContextEstimation} incorporated multi-context attention by utilizing CRFs to model the correlations between neighbouring regions in the attention map. \emph{Deeply-Learned-Compositional-Model or DLCM}~\cite{TangDeeplyEstimation} used SHG as their backbone and employed DNNs to learn the compositionality of the human-body by enforcing a bone-based part representation as the output of intermediate stacks. The learned compositionality improved the performance significantly and outperformed all other contemporary methods.

In order to present a comprehensive analysis of the effects of adversarial attack on 2D HPE systems, we carefully chose a representative set of HPE systems that span direct-regression, heatmap and anthropometric-employment based approaches. Our selection ensures at least one representative approach pertaining to major architectural and other choices that yields competitive results on the MPII \cite{andriluka14cvpr}, a benchmark for 2D HPE systems. Below we provide the details of the set of HPE systems that we analyzed from the perspective of adversarial robustness. Whenever possible we use the released networks from the authors, otherwise we implement it ourselves and make sure we get with in $5\%$ of their original reported accuracy.
\newline
\textbf{DeepPose~\cite{ToshevDeepPose:Networks}:} It's a direct-regression approach that uses an AlexNet backbone and regresses for the pixel coordinates in image space. Since there was no official code or pre-trained models we implemented it with an Image-Net pre-trained ResNet-34 backbone and fine-tuned for both MPII and COCO datasets. Note that we do not use multi-stage feedback used in original paper but still achieve comparable results to the original model. 
\newline
\textbf{Stacked Hourglasses~\cite{NewellStackedEstimation}:} It's a heatmap-based approach that employs an hourglass-like encoder-decoder structure, which consists of a sequence of convolution layers followed by up-sampling layers with skip connections. Each pair of up/down-sampling layers are referred to as a stack, and the previously predicted heatmaps are concatenated with the visual features and input the next stack. Typically, they are referenced based on the number of stacks, $s$, as s-SHG. E.g. a two stack hourglass backbone would be referred to as 2-SHG. In the original paper, the authors used an 8-SHG architecture. In order to clearly bring out the effect of the number of stacks w.r.t. adversarial attacks, we implemented 2-SHG and 8-SHG architectgures. For the 8-SHG architecture, we use the official code-base and MPII pre-trained models provided by the authors. For the 2-SHG, we train the models ourselves on both MPII \& COCO dataset.
\newline
\textbf{Chained Predictions Network~\cite{GkioxariChainedNetworks}:} It's a heatmap-based approach with anthropometric information, It predicts the body-joints in a sequential manner where the next joint's location is conditioned on the previously predicted joints. Intuitively, it aims at modeling the anthropometric structure as a sequence prediction. It employs ImageNet pre-trained backbone followed by a series of convolution-deconvolution pairs to generate heatmap for one joint at a time with the previously predicted heatmaps concatenated with the visual features. Due to the lack of official code-base of pre-trained model, we implemented it with a ResNet-34 backbone with deception layers (multi-scale de-convolution) as described in the original paper. This system is chosen to reflect the effects of adversarial attack on cascased prediction.
\newline
\textbf{Pose Attention~\cite{ChuMulti-ContextEstimation}:} It's yet another heatmap-based approach that uses SHG backbone and employs CRFs for capturing the anthropometric structure in the form of correlations within the heatmaps. Additionally, it also introduced a novel Hourglass Residual Module with larger kernels to afford larger receptive fields. We use the pretrained model provided by the authors for evaluation on MPII dataset.
\newline
\textbf{Deeply-Learned-Compositional-Model or DLCM~\cite{TangDeeplyEstimation}:} This approach is also a heatmap-based approach that explicitly learns the compositionality of human bodies. In addition to localizing the joints, it learns the high-order relationships among body parts as well. It uses a 5-SHG architecture as the backbone with the first and the last stack regress for joints, second and penultimate stack regress for bones and the third stack corresponding to higher-order relations. We used the pre-trained models provided by the authors for our analysis.

\subsection{Multi-Person 2D Human-Pose Estimation Systems (MHPE)}
While single-person 2D HPE systems are at the heart of the research in the community, practical 2D-HPE systems are inherently multi-person in nature i.e. they need to work on images/videos that contain multiple persons. Such systems require to correctly identify all the joint locations for a variable number of humans in an input image. Recent approaches for multi-person 2D-HPE systems \cite{openpose, cascaded_pyramid, bottom_up_hrnet, top_down_hrnet} can be broadly categorized into a) bottom-up and b) top-down approaches. Since, the focus of our analysis is single-person 2D-HPE systems, we only consider the state-of-the-art systems for multi-person 2D-HPE systems and leave a thorough analysis of such systems for future work. 
\subsubsection{Top-Down Multi-Person 2D HPE Systems}
As the name suggests, this approach first detects all the humans bounding-boxes in the image and then predict the pose for each of them. Typically, top-down approaches are more accurate than the bottom-up approaches but they are comparatively slower due to multiple forward passes of model. The higher accuracy could be attributed to merging the state-of-the-art advancements in person-detection and single-person 2D-HPE systems. We select the HR-Net Pose-Estimation \cite{top_down_hrnet} as a representative top-down approach for analysis owing to it's state-of-the-art performance among top-down methods. It uses a pretrained Faster RCNN-50 \cite{faster_rcnn} for person-detection and HR-Net backbone for pose estimation. Since, we already carry out a thorough analysis of single-person 2D-HPE systems, we focus on attacking the object detection module to illustrate the effect of adversarial attacks on multi-person 2D-HPE systems. Furthermore, end-to-end attack on both the modules, detection and 2D-HPE, isn't a trivial problem due to the presence of a non-differentiable cropping and resizing step, which we leave for future work to address. While attacking only the object detector gives us the lower bound on the degradation caused due to adversarial attack, it can still serve to provide some useful insights. Technically, for an input image $I$, we perform an adversarial attack on the object detection network to obtain the modified image $I'$. We then use $I'$ as the input for both the object detection network and the subsequent 2D-HPE system.
\subsubsection{Bottom-Up Multi-Person 2D-HPE Systems}
Unlike top-down approaches, bottom-up methods predict all humans in a single shot by predicting a single heatmap per joint, followed by grouping operations to associate each predicted joint to distinct individuals in the image. The association can be performed using 1.) Part Affinity Fields \cite{openpose} that model 2D vector-fields over image domain while encoding the location and orientation of limbs to facilitate the grouping of different joints together, or 2.) Associative Embeddings \cite{associative_embedding} that model tag-embeddings corresponding to every predicted joint and aims at grouping the humans based on the L2-distance between tag representations assigned to different joints. For our analysis, we select the state-of-the-art Higher-HR-Net model \cite{bottom_up_hrnet}, which employs associative embeddings for bottom-up approach. In addition, \cite{bottom_up_hrnet} and \cite{top_down_hrnet} share the same HR-Net backbones while only differing in their approach towards multi-person setting, thereby, afford a close comparison between the top-down and bottom-up approaches. The network input is an image with multiple humans and it predicts heatmaps for joints and associative embeddings, followed by a matching step and finally returning the multi-person skeletons. For the bottom-up approaches, we attack using both the heatmap losses and associative embedding losses to ascertain its robustness against adversarial attacks.

\subsection{Single-Person 3D Human-Pose Estimation Systems}
While 2D-HPE systems are the workhorse for multi-media analysis, an immersive and real-world human-compute-interaction also requires accurate 3D-HPE systems, which are needed for estimating the human-pose in 3D coordinates. The prediction of exact 3D coordinates from a monocular image is an ill-defined and under-constrained problem because there could be multiple 3D-skeletons for the same 2D projection, therefore, a popular work-around is to regress for the relative depth of the 2D-joints. There exist various approaches for single-person 3D-HPE systems \cite{Zhou_2017_ICCV, Dabral_2018_ECCV, Hossain_2018_ECCV, iskakov2019learnable, nibali2018margipose, Wandt2019RepNet, zhaoCVPR19semantic, videopose3d2019,fabbri2020compressed,pavlakox3dpose} , but a thorough analysis of such systems is beyond the scope of this article, therefore, we only analyse the popular system presented in \cite{Zhou_2017_ICCV} with an aim to depict the generality of our results from 2D-HPE systems. The approach uses a SHG architecture to first predict the 2D-keypoints, followed by a direct regression of the relative depth coordinates for each joint. We use this architecture owing to its simplicity, use of direct-reression and popularity (\cite{Hossain_2018_ECCV}, \cite{Dabral_2018_ECCV} built on top of this work). The Human3.6M dataset \cite{h36m_pami} is used with Mean Per Joint Error (MPJPE) as the evaluation metric. Our pretrained model achieves 60 MPJPE score on this dataset (lower is better).  

\subsection{Adversarial Attack}\label{subsubsec:AdversarialAttack}
In this section, we provide a brief overview of the fundamentals of adversarial attacks, their types and our proposed HPE-specific adversarial attack schemes to facilitate their basic understanding and the effects of involved hyper-parameters. For a detailed technical understanding, we request the reader to go through the references. Adversarial attacks consist of modifying the original image, $I$, with imperceptible changes to the human eye with an aim to significantly alter the output of a DNN network $y = f(I;\theta)$. Typically, it's achieved by corrupting $I$ with an additive adversarial noise, $n$, to yield $I_n$. The core mechanism behind adversarial noise generation is back-propagating the error signals for a corrupted output, through the network, up to the input image. The back-propagated error at the image is the adversarial noise $n$. In order to keep the noise visually imperceptible, the pixel-wise magnitude of the noise, $n$, is constrained to be smaller than a pre-defined threshold, $\epsilon$, i.e. $||n||_{\infty} < \epsilon$. The adversarial attacks can be categorized based on the availability of the network and/or the input image, the adversarial corruption target, and the process of obtaining the adversarial noise. Different combinations of these choices give rise to different flavors of adversarial attacks, but the core mechanism of back-propagating error signal to the image remains the same.

\begin{figure*}[!t]
\centering
\includegraphics[width=1.0\linewidth ]{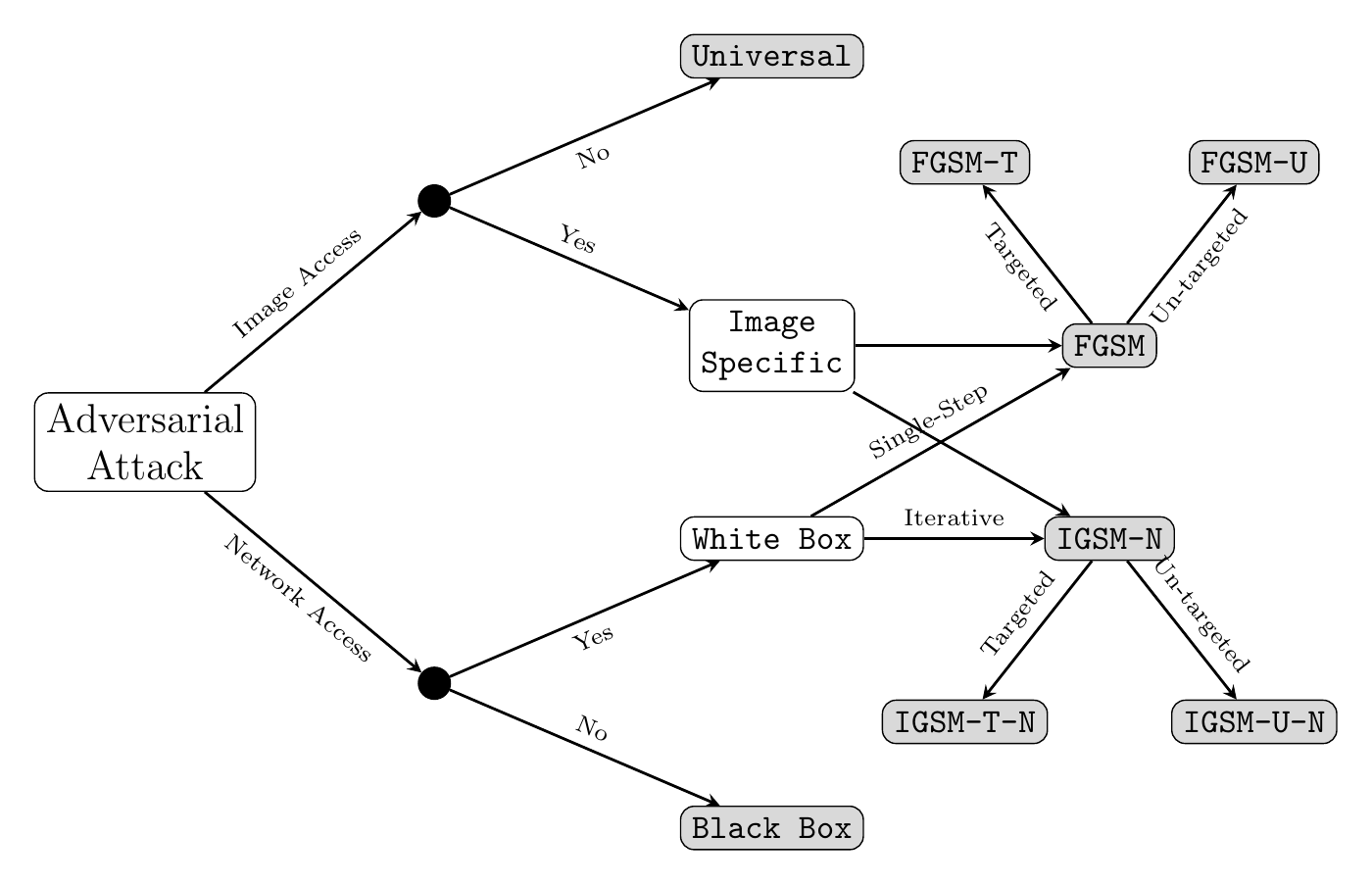}

\caption{Overview of different adversarial perturbation schemes w.r.t. the access to the network/image, targeted or un-targeted, iterative or single-step and image-agnostic or image-specific. The gray-color boxes show the specific instances of these combinations.}  
\label{fig:adv_attack}
\end{figure*}

As reviewed in the Related Work section, there are multiple approaches for back-propagation based noise generation. Among these approaches, the Fast Gradient Sign Method, or FGSM,~\cite{GoodfellowEXPLAININGEXAMPLES} is the most popular and computationally efficient while being equally effective. FGSM explicitly bounds the $l_{\infty}$ norm of every pixel by using the scaled, by $\epsilon$, sign of gradient w.r.t. the desired objective to obtain $n$. Fig.~\ref{fig:adv_attack} shows different adversarial perturbation combinations for a quick overview and describe their details. We start by describing the simplest case where both the network and the image are available for generating the perturbation and draw a distinction between un-targeted and targeted attacks, followed by their iterative counterparts. Then we describe the difference between image-specific and universal attacks followed by white-box and black-box attacks that depends on the availability of the network. Please note that the aforementioned order is different from the hierarcy shown in Fig.~\ref{fig:adv_attack}; it's chosen for the ease of understanding.

\subsubsection{Targeted vs. Un-targeted Perturbations}\label{sssec:untar_vs_tar}
The attacks could either be un-targeted or targeted towards a desired target output. Un-targeted attacks simply try to increase the loss of the network for a given pair of input and label $(I, y)$ to obtain the perturbed image $I^p$ as-
\begin{equation} \label{eq:untargetFGSM}
    I^p = I + \epsilon.sign(\nabla_I \mathcal{L}(f(I;\theta),y))
\end{equation}
On the other hand, the targeted attack tries to push the output of the network towards a desired $y^t$. For classification systems, $y^t$ can be easily obtained as the least likely, from domain knowledge, or the desired class ~\cite{KurakinADVERSARIALSCALE}. Unfortunately, there is no counterpart of least likely pose for HPE systems. Therefore, we propose to choose a target pose, $P^t$ from the pool of ground-truth poses from the validation set, $\mathcal{P} = \{\hat{P}_1, \hat{P}_2,\dots \}$, for which the PCKh is equal to 0 w.r.t. the input image $I$. Intuitively, this is akin to selecting the most \emph{unlikely} pose for $I$. Please note that due to the nature of PCKh loss \cite{andriluka14cvpr}, there could be multiple poses that satisfy PCKh equal to 0 for $I$, hence, we randomly select one of them for our analysis and generate the perturbed Image $I^p$ as-
\begin{equation} \label{eq:targetedFGSM}
    I^p = I - \epsilon.sign(\nabla_I \mathcal{L}(f(I;\theta),P^t))
\end{equation}
The un-targeted and targeted attacks are referred to as \textbf{FGSM-U} and {\bf FGSM-T}, respectively.

\subsubsection{Single-Step vs. Iterative Perturbations}\label{sssec:iterative_attack}
Both, FGSM-U and FGSM-T, can be extended to their iterative counterparts \textbf{IGSM-U-N} and \textbf{IGSM-T-N}, respectively, that take {\bf N} iterations to yield the final perturbed image $I^p$ starting with $I$. The perturbed image $I^p_i$ for the $i^{th}$ iteration for un-targeted (Eq.~\ref{subeq:ituntarFGSM}) and targeted (Eq.~\ref{subeq:ittarFGSM}) attack is given as-
\begin{subequations}
\begin{align}
    &I^p_i = C_{\epsilon}(I, I^p_{i-1} + \alpha.sign(\nabla_{I^p_{i-1}} \mathcal{L}(f(I^p_{i-1};\theta),y)))\label{subeq:ituntarFGSM}\\
    &I^p_i = C_{\epsilon}(I, I^p_{i-1} - \alpha.sign(\nabla_{I^p_{i-1}} \mathcal{L}(f(I^p_{i-1};\theta),P^t)))\label{subeq:ittarFGSM}\\
    &s.t.\ x_0-\epsilon \leq C_{\epsilon}(x_0, x_i) \leq x_0+\epsilon
\end{align}
\end{subequations}
where, $C_\epsilon(x_0, x)$ clips $x$ to $[x_0-\epsilon, x_0+\epsilon]$. The iterative attacks are typically more detrimental as compared to single-step attacks because they can corrupt the image in a highly non-linear fashion owing to multiple iterations. 

\subsubsection{Image-Specific vs. Universal Perturbation}\label{sssec:universal_perturbation}
So far, the aforementioned discussed approaches for obtaining the perturbed image, $I^p$, are image-specific and require access to the initial input image and employ expensive back-propagation steps to obtain $I^p$ (Eqn.~\ref{eq:untargetFGSM}, \ref{eq:targetedFGSM}, \ref{subeq:ituntarFGSM}, \ref{subeq:ittarFGSM}). In order to get rid of the costly back-propagation steps at the time of generating perturbed image ~\cite{Moosavi_DezfooliUniversalPerturbations} show that it's possible to learn an image agnostic or \emph{universal perturbations} from a dataset that can generalize to unseen images. The universal perturbation is obtained by optimizing for a perturbation that can maximally degrade the performance for a representative set of images. Practically, universal perturbations are computed by iterating over the dataset or a subset of dataset and aggregating the individual perturbations. For our analysis, we adopt the method presented in~\cite{Metzen_2017_ICCV} to HPE setting and obtain the universal perturbation $u$ by computing the perturbations on training samples $x_i$, or mini-batches of them, and aggregating them to obtain the final $u$ after re-scaling- 
\vspace{-0.5em}
\begin{equation}
    u = u + \delta.sign(\nabla_{x_i}\mathcal{L}(f(x_i;\theta),y)
\end{equation}
We fix $\delta = \frac{\epsilon}{200}$, mini-batch size of 16 and $\|u\|_{\infty} \in \{8, 16\}$, because lower $\epsilon$ values hindered learning while higher values are perceptible and use the same setting for all the architectures. The obtained $u$ can be simply added to any image to attack the network, therefore, making it more widely applicable than network access attacks.

\subsubsection{White-box vs. Black-box Perturbations}\label{sssec:blackbox_attack}
All the aforementioned approaches require complete access to the network that they are trying to attack, which is not practical from the perspective of real-world systems because network access can easily be controlled by hardware-level encryption. Therefore, \cite{paperblack} proposed to employ a \emph{source} network and learn adversarial perturbations to attack a \emph{target} network. Surprisingly, they report that even without the access of the target network, except while evaluating the performance, the aforementioned \emph{black-box} perturbations, learned only from the source network, were significantly degrading the performance of the target network. This phenomenon indicates that different networks are leveraging similar low-level image details during learning and this can also serve as a possible venue for adversarial attack. Such black-box perturbations can either be image-specific, obtained by FGSM-U/T or IGSM-U/T, or image-agnostic universal perturbations. The latter gives rise to \emph{doubly black-box} attacks i.e. we need neither access to the target network nor do we need the image to obtain the perturbation; the most detrimental of all the attacks from the perspective of real-world systems. 

\subsection{Evaluation Protocol and Dataset}\label{ssec:evalprotocol}
In order to ensure uniformity across all the HPE systems, we use a standard protocol to evaluate the performance on the validation sets that includes similar cropping and data pre-processing. Therefore, in some cases our reported results are a little inferior to the originally reported results that employ flipping, multiple crops and other augmentation techniques to boost their performance. We analyzed the selected HPE systems on two different pose databases - MPII \cite{andriluka14cvpr} and COCO \cite{coco} - to show the generalizability of our findings. All the results are reported on the validation set and we use the standard PCKh \cite{andriluka14cvpr} and OKS \cite{oks} metrics to measure the pose-estimation performance for MPII and MS-COCO. Assuming that the Euclidean distance between the predicted keypoint and its ground-truth location and the visibility of the keypoint are denoted by $d_i$ and $v_i$ respectively, $i \in \{1,2 ... k\}$, PCKh is computed as $\frac{\sum_{i = 1}^k \delta(d_i \leq 0.5 * h) \cdot \delta(v_i > 0) }{\sum_{i = 1}^k \delta(v_i > 0)}$ where $h$ is the head size of a person and $\delta(x)$ is a function that evaluates to 1 if $x$ is true, else it evaluates to 0. The result is a binary score assigned to each joint which is then averaged across all visible joints. The OKS metric, on the other hand, provides a score lying in the continuous region $[0,1]$ using the formula: $\frac{\sum^{k}_{i=1}e^{\frac{d_i^2}{2s^2k_i^2}}\delta(v_i > 0) }{\sum_{i = 1}^k \delta(v_i > 0)}$


Since the initial performance of the analyzed HPE systems differs, it's not fair to compare the degradation due to adversarial attacks by comparing the drop in absolute performance. Therefore, for un-targeted and universal attacks, we report relative-PCKh given by $(perturbed~performance/original~performance)*100$ score-ratio where lower values indicate lower robustness against adversarial attack. For the targeted attacks, on the other hand, we report the targeted-PCKh or absolute-PCKh of the output w.r.t. to the adversarial target, therefore, higher values indicate lower robustness against adversarial attacks. The strength of adversarial attack is measured in terms of $||I - I^p||_\infty \leq \epsilon$, where $\epsilon \in \{0.25, 0.5, 1, 2, 4, 8, 16, 32\}$, hence, higher values of $\epsilon$ indicate more aggressive adversarial attacks. For iterative attacks, IGSM-U/T-N, the strength of the attack increases with the number of iterations $N$. The popular setting for classification systems is $N= 10$, but our experiments indicate that the HPE systems are relatively more robust, therefore, we also report the result with 100 iterations or $N \in \{10, 100\}$. We also note that the targeted attacks are more difficult that un-targeted attacks, \ref{subsubsec:targetedvs}, therefore, for targeted attacks we used 20 iterations instead of 10. Overall, it yields four different configurations of attacks: \textbf{IGSM-U-10}, \textbf{IGSM-T-20}, \textbf{IGSM-U-100}  and \textbf{IGSM-T-100}. For the iterative attacks, we observe that the optimal value of the step-size, $\alpha$, falls in the range $[\frac{\epsilon}{3},\frac{\epsilon}{2}]$ for un-targeted and in $[\frac{\epsilon}{9},\frac{\epsilon}{7}]$ for targeted attacks. We report the results of IGSM-U/T-100 with the popular setting of $\epsilon = 8$ and refer to the Appendix for the results with other values of $\epsilon$, while IGSM-U/T-10/20 results are reported for all $\epsilon$ values.

\begin{figure*}[!th]

\centering
\includegraphics[width=1.0\linewidth ]{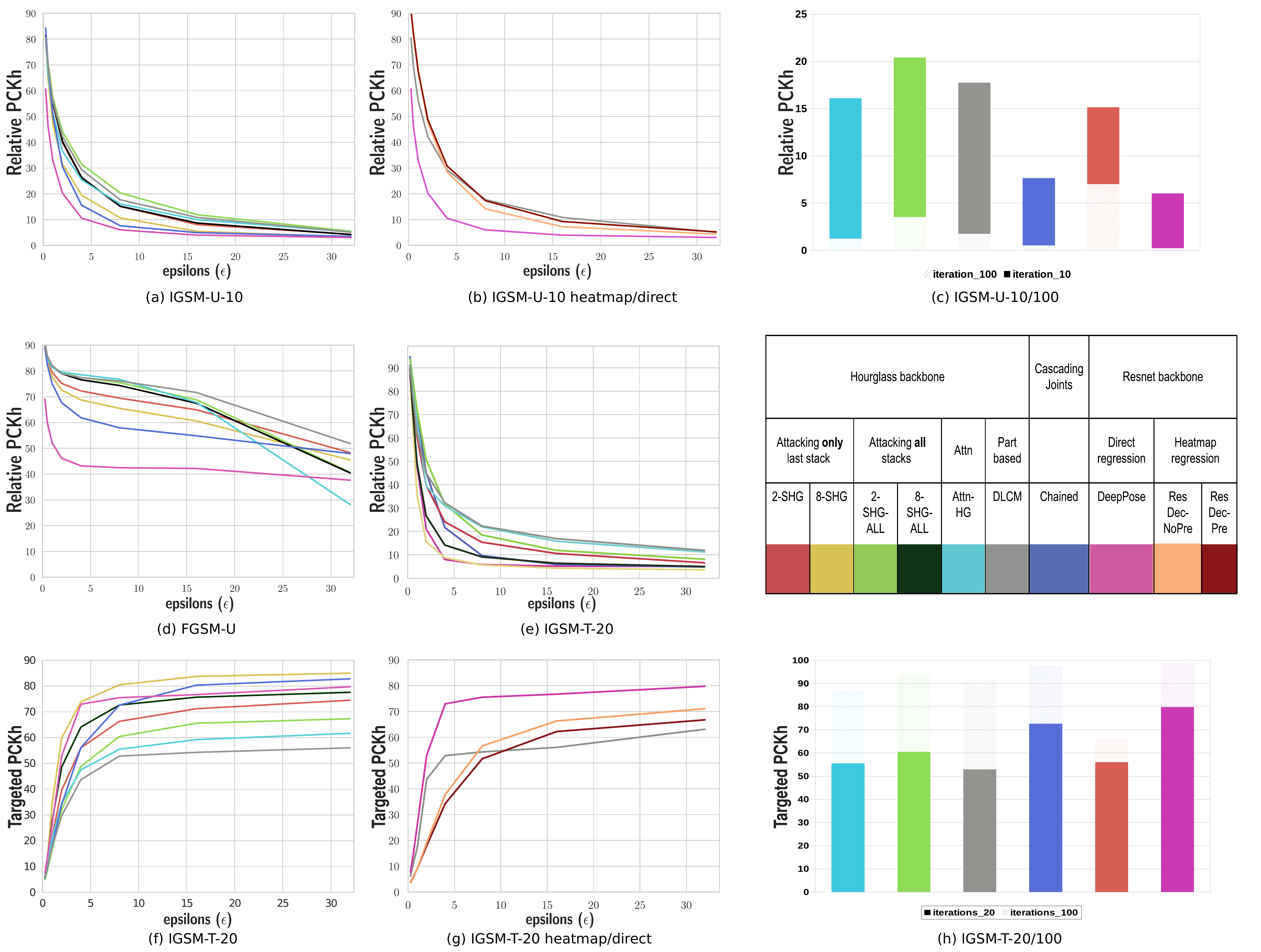}
\caption{Performance of all the models under different types of attacks. First two rows contain relative-PCKh measured under different kinds of attacks including FGSM-U, IGSM-U-10, IGSM-U-100, IGSM-T-20. Last row measures absolute-PCKh/targeted-PCKh of model in targeted attack setting under IGSM-T-10 and IGSM-T-100 attacks.\\(a) depicts the relative-PCKh as a function of $\epsilon$ for IGSM-U-10 attack \\\hspace{\textwidth} (b) depicts relative-PCKh for heatmap vs regression models under IGSM-U-10 attack  \\\hspace{\textwidth} (c) depicts performance under IGSM-U-10 vs IGSM-U-100 attack. \\\hspace{\textwidth} (d) depicts relative-PCKh as function of $\epsilon$ for FGSM-U attack \\\hspace{\textwidth} (e) depicts relative-PCKh as function of $\epsilon$ for IGSM-T-20 attack \\\hspace{\textwidth} (f) depicts the absolute-PCKh as a function of $\epsilon$ for IGSM-T-20 attack \\\hspace{\textwidth} (g) depicts absolute-PCKh for heatmap vs regression models under IGSM-T-20 attack \\\hspace{\textwidth} (h) depicts performance under IGSM-T-20 vs IGSM-T-100 attack.}

\label{fig:allplots}

\end{figure*}

\section{Analysis of Adversarial Attack on 2D HPE Systems}\label{sec:experiment_analysis}
In this section, we once again follow the order in which we described different schemes for adversarial attacks, Sec.~\ref{subsubsec:AdversarialAttack}, to study the effects of different choices on adversarial attacks for HPE systems. Specifically, we first contrast the effects of un-targeted and targeted attacks followed by the effect of iterative attack vs single-step attacks. We then contrast the effects of image-specific vs. universal attacks followed by a discussion on white-box vs. black-box attacks. Next, we include a comparison between HPE systems and other systems, like classification, semantic segmentation, to conclude that HPE systems are relatively more robust. We then compare different 2D HPE systems in terms of their robustness against adversarial attacks followed by an interesting analysis of the robustness of different body-joints against adversarial attack. Lastly, we show additional analysis on the effects of adversarial attack on 2D multi-person HPE and single-person 3D HPE systems as well. Such a wide array of analyses gives rise to multiple tables and graphs, which can hide the interesting insights that are potentially more useful and interesting than the numbers themselves. Therefore, we focus more on the message and transfer the tables and graphs to the Appendix, wherever possible. In the spirit of the aforementioned strategy, we show the results of our analyses on MPII dataset \cite{andriluka14cvpr} in the main manuscript and transferred the results on COCO dataset ~\cite{coco}  to the Appendix.

\subsection{Targeted vs. Un-targeted Attacks} \label{subsubsec:targetedvs}
Here we are interested in understanding the difference between the effects of un-targeted vs targeted attacks on HPE systems. First, we note that the targeted attacks are more difficult to execute than un-targeted because they require more iterations, 20 vs. 10, therefore, we couldn't execute FGSM-T, unlike FGSM-U, and had to resort to the iterative version i.e. IGSM-T-20/100. For the targeted attacks, we consider the absolute-PCKh achieved on target pose as a measure of effectiveness. Additionally, for IGSM-T-20, we also compute the relative-PCKh w.r.t the original pose to obtain a relative score. Note that since we choose the target poses such that the PCKh between the original and target pose is 0, therefore, the relative-PCKh can potentially fall to 0 due from targeted attacks as well. Comparing the relative-PCKh obtained for IGSM-U-10, $\sim5$ (Fig.~\ref{fig:allplots}(a)) vs. relative-PCKh for IGSM-T-20, $\sim10$ (Fig.~\ref{fig:allplots}(e)), reveals that targeted attacks are weaker than un-targeted. Intuitively it makes sense because an un-targeted attack can take large steps in the direction of increasing loss while the targeted attack requires finding the optimal $I^p: \|I - I^p\|_{\infty} <= \epsilon$ where the loss $\mathcal{L}(f(I^p;\theta),P^t)$ is small; evidently a more difficult problem. This also explains the fact that optimal value of the step-size, $\alpha$, for IGSM-T is $\sim 3$ times smaller than that of IGSM-U, which is needed for driving the more complex objective. Sufficient number of such small step-size iterations, $\sim 100$, of targeted attacks can still lead to almost 100\% target PCKh \ref{fig:allplots}(h). Yet another interesting difference between un-targeted and targeted attacks can be observed by contrasting the behaviour of different HPE systems for higher values of $\epsilon$, i.e. stronger attacks, from Fig.~\ref{fig:allplots}(a) and Fig.~\ref{fig:allplots}(f). For un-targeted attack, different HPE systems converge to in their adverse effects, on the other hand, they diverge under targeted attack! It indicates that perhaps under extreme targeted attack different networks perform significantly different in terms of their robustness.

\subsection{Effect of the Number of Iterations on the Attack}\label{ssec:exp_iterative}
In this sub-section, we study the effect of number of iterations of an iterative attacks (IGSM-U/T-N) on HPE systems. First we compare the relative drop in PCKh for FGSM-U and IGSM-U-10 with the help of Fig. \ref{fig:allplots}(a) and \ref{fig:allplots}(d), respectively. Clearly, the iterative attacks are more effective than single-step attacks, for example- with $\epsilon = 8$, even the least effected HPE system (2-SHG-ALL) has a relative-PCKh $\sim20$ for IGSM-U-10 vs. $\sim75$ for FGSM-U. Moreover, the adverse effects only increase with further increase in the number of iterations; Fig.~\ref{fig:allplots}(c) and Fig.~\ref{fig:allplots}(h) show the dramatic degradation across all the HPE systems by comparing IGSM-U-10 vs. IGSM-U-100 and IGSM-T-10 vs. IGSM-T-100, respectively. Specifically, IGSM-U-100 drives the relative-PCKh to less than 5 and IGSM-T-100 achieve targeted-PCKh $\sim90$ even at $\epsilon = 8$. This observation is in stark contrast with the effect of IGSMs on classification or semantic segmentation problems, where ~\cite{Kurakin2017ADVERSARIALWORLD} reported that $min(\lceil{1.25 \epsilon}\rceil, \epsilon + 4)$ iterations are sufficient for complete degradation. On the other hand, HPE systems often need up to 100 iterations for the same. Unfortunately, however, with enough iterations, all the systems degrade by over 95\% which shows that all models are vulnerable for carefully designed perturbations. See appendix Sec. \ref{hundrediter} for results on all $\epsilon$ values.

\subsection{Image-Agnostic Universal Adversarial Perturbations}\label{ssec:exp_universal}
\begin{figure}[b]
\vspace{-1em}
\centering

    \subfloat[Attention Hourglass]{
        \includegraphics[width=0.47\linewidth]{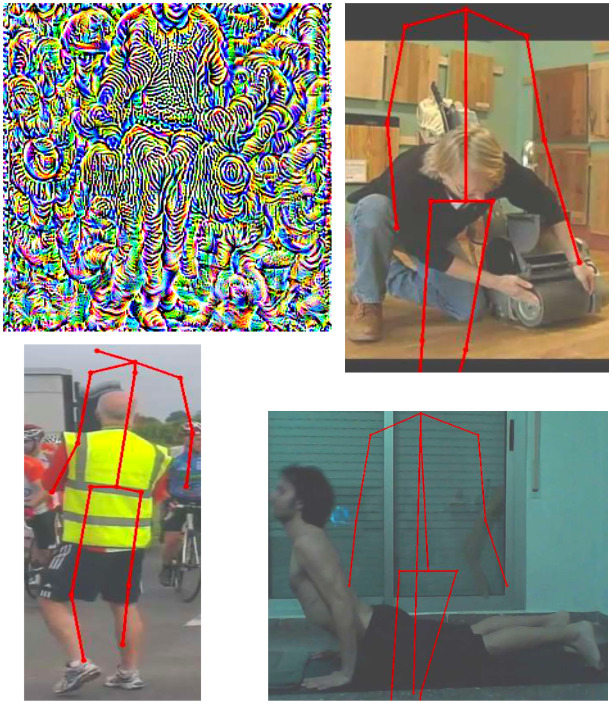}}
    \hfill
    \subfloat[DLCM]{
        \includegraphics[width=0.47\linewidth]{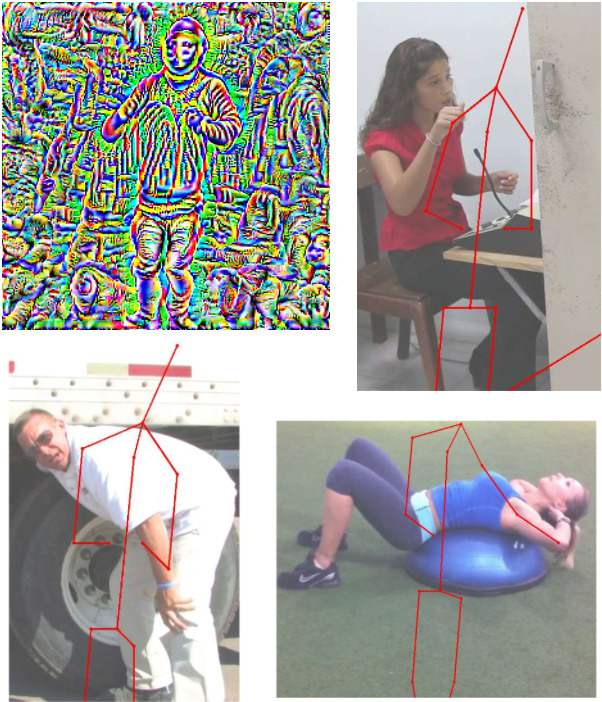}}
    \hfill
    
\caption{Examples of predictions after adding image-agnostic universal perturbations along with the corresponding scaled perturbation. For all the images, the initial predictions made by the networks were correct.}
\label{fig:universal_examples}
\end{figure}

Up till now the analysis has focused on image-specific adversarial attacks, in this section we analyze the effects of universal perturbations on HPE systems. We follow Sec.~\ref{sssec:universal_perturbation} to obtain the universal adversarial perturbations for all the considered architectures. Once obtained, they can be simply added to any input image to fool the corresponding architecture, making them practically useful in real-world scenario. The degradation in the performance of different HPE system using universal perturbations with $\epsilon = 16$ (see Appendix for $\epsilon=8$ results) are shown in Table~\ref{table:blackBox} (the underscored diagonal entries). Averaged over all the HPE systems, universal attacks degrade the PCKh values on training (used to obtain them in the first place) and validation sets to 6.4\% and 9.9\% of their original value, respectively. It clearly demonstrates that even image-agnostic attacks can render all the HPE systems practically useless. Moreover, universal attacks' effect is similar to image-specific iterative attacks, 9.9\% vs. $\sim8\%$, for $\epsilon=16$ (see Fig~\ref{fig:allplots}). In order to ascertain the effect of the amount of training data required for obtaining universal perturbations, we obtained them with varying number of samples from the training set, as in ~\cite{Moosavi_DezfooliUniversalPerturbations}. We observe that even with 10\% data samples, i.e. only 2500 images, the obtained universal perturbations degrade the performance to 18\% vs. 9.9\% with all the 25925 samples. Therefore, we conclude that universal perturbations are extremely effective and compute efficient attack mechanism.

\begin{figure*}[b]
\vspace{-1em}
\centering

    \subfloat[2-Stacked Hourglass  \label{fig:uni_2shg}]{
        \includegraphics[width=0.30\linewidth]{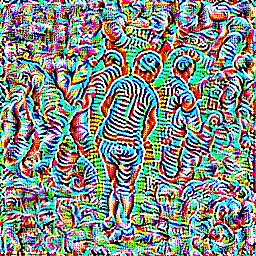}}
    \hfill
    \subfloat[Chained Predictions  \label{fig:uni_chained}]{
        \includegraphics[width=0.30\linewidth]{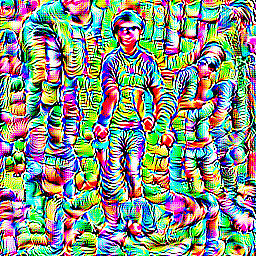}}
    \hfill
    \subfloat[Attention Hourglass  \label{fig:uni_attention}]{
        \includegraphics[width=0.30\linewidth]{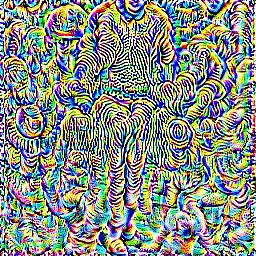}}
    \\
    
\vspace{-0.025\linewidth}
    
    \subfloat[8-Stacked Hourglass  \label{fig:uni_8shg}]{
        \includegraphics[width=0.30\linewidth]{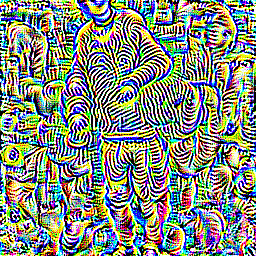}}
    \hfill
    \subfloat[DLCM  \label{fig:uni_DLCM}]{
        \includegraphics[width=0.30\linewidth]{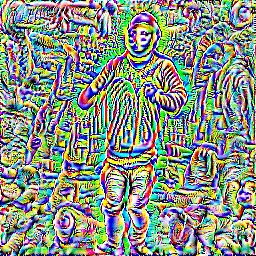}}
    \hfill
    \subfloat[DeepPose  \label{fig:uni_deeppose}]{
        \includegraphics[width=0.30\linewidth]{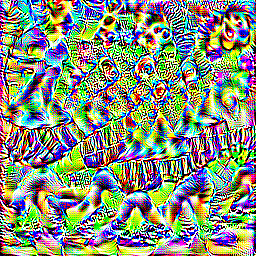}}

\caption{Visualization of image-agnostic universal perturbations, with $\epsilon=8$, for different networks scaled between 0 to 255 for better visualization. Note the hallucinated body-joints, mostly arms and limbs to fool HPE networks.}
\label{fig:UniversalAdversarial}
\end{figure*}

In order to reveal the inner workings of the universal attacks, we plot the universal perturbations for $\epsilon = 8$, scaled between 0 to 255 for better visualization, obtained for different HPE systems in Fig.~\ref{fig:UniversalAdversarial}, more such visualization are shown in the Appendix Fig.~\ref{fig:UniversalAdversarial_eps8}, \ref{fig:UniversalAdversarial_small}. To the best of our knowledge, it's the first such visualization of adversarial perturbations for HPE that clearly reveals it's working via human-body hallucinations. A closer look reveals that universal perturbations attack HPE systems by hallucinating body-joints, mostly limbs, throughout the image. Lastly, we also show that often the predictions on corrupted images are similar to the hallucinated poses for different images despite the fact that these perturbations were never explicitly designed to predict such specific outputs. See Fig.~\ref{fig:universal_examples} (\& Fig. \ref{univpdfs_pyranet},\ref{univpdfs_newell},\ref{univpdfs_attention},\ref{univpdfs_shg}, \ref{univpdfs_chained},\ref{univpdfs_deeppose} in appendix) for a few such cases where the original predictions of the network changed to a similar incorrect pose for all the images. Moreover, these skeletal predictions resemble humans centered in the perturbation. It is interesting to note that while the visualizations of universal perturbations hallucinate the human body, the visualization of such perturbation for DeepPose does not! It could be due to the difference in the loss function between DeepPose and heatmap-based approach, where the former directly regresses for joints while the latter explicitly searches for the body-joint locations. 

\begin{table*}[!ht] 
\centering

 \begin{tabular}{c | c c c c c c c c c c} 
 
 \hline
 &   & \hspace{-7px}  8-SHG & \hspace{-7px}  8-SHG- & \hspace{-7px}  Attn- &   DLCM & \hspace{-7px}  2-SHG- & \hspace{-7px}  2-SHG & \hspace{-7px}  Chained & \hspace{-7px}  Deep- & \hspace{-7px} Doubly\\ 
 & & \hspace{-7px}   & \hspace{-7px}  ALL & \hspace{-7px}  HG & \hspace{-7px}  & \hspace{-7px}  ALL & \hspace{-7px}  & \hspace{-7px}   & \hspace{-7px}  Pose & \hspace{-7px} \\ 
 [0.5ex]
 \hline
 \hline
 
\multirow{6}{*}{\rotatebox[origin=l]{90}{\hspace{-3px} \textbf{Target Netwo\vspace{-12pt}rk}}} & \hspace{-7px} 8-SHG & \hspace{-7px} \underline{8.85} & \hspace{-7px} \underline{5.92} & \hspace{-7px} \textbf{53.32} & \hspace{-7px} 56.61 & \hspace{-7px} 53.45 & \hspace{-7px} 68.17 & \hspace{-7px} 63.23 & \hspace{-7px} 86.7 & \hspace{-7px}  63.58\\
 &  Attn-HG & \hspace{-7px} \textbf{41.92} & \hspace{-7px} 48.47 & \hspace{-7px} \underline{11.47} & \hspace{-7px} 57.62 & \hspace{-7px} 61.05 & \hspace{-7px} 71.68 & \hspace{-7px} 68.1 & \hspace{-7px} 84.78 & \hspace{-7px}  61.95\\
 &  DLCM & \hspace{-7px} \textbf{46.76} & \hspace{-7px} 47.09 & \hspace{-7px} 60.07 & \hspace{-7px} \underline{12.75} & \hspace{-7px} 64.45 & \hspace{-7px} 74.02 & \hspace{-7px} 67.41 & \hspace{-7px} 84.93 & \hspace{-7px}  63.53\\
 &  2-SHG & \hspace{-7px} 51.95 & \hspace{-7px} 55.17 & \hspace{-7px} 75.28 & \hspace{-7px} 70.08 & \hspace{-7px} \underline{10.35} & \hspace{-7px} \underline{15.7} & \hspace{-7px} \textbf{51.6} & \hspace{-7px} 88.59 & \hspace{-7px}  65.45\\
 &  Chained & \hspace{-7px} 77.65 & \hspace{-7px} 79.7 & \hspace{-7px} 82.57 & \hspace{-7px} 81.15 & \hspace{-7px} \textbf{72.08} & \hspace{-7px} 78.45 & \hspace{-7px} \underline{10.96} & \hspace{-7px} 75.36 & \hspace{-7px}  78.14\\
 &  DeepPose & \hspace{-7px} 74.19 & \hspace{-7px} 70.44 & \hspace{-7px} 75.12 & \hspace{-7px} 75.03 & \hspace{-7px} 72.23 & \hspace{-7px} 75.6 & \hspace{-7px} \textbf{42.04} & \hspace{-7px} \underline{2.78} & \hspace{-7px}  69.24\\
 \hline
 \end{tabular}
 \vspace{-5pt}

\caption {The results of all source \& target pairs under doubly black-box attack setting (UAP generated from training dataset tested on validation dataset). Rows represent the relative degradation in the target network when attacked by the network in the column. \textbf{Doubly} stands for the relative drop in performance from a doubly-black-box attack. \textbf{Boldface} shows the strongest black box attack for a model and \underline{underlined} numbers indicate the performance of the model on itself}
\vspace{-10pt}
\label{table:blackBox}
\end{table*}

\subsection{Black-Box Attacks}\label{ssec:exp_black_box}
Following the definition of black-box attacks from Sec.~\ref{sssec:blackbox_attack}, we report all combinations of \emph{source} and \emph{target} network pairs $(S \longrightarrow T)$ and tabulate the results of \emph{doubly black-box} attacks in Table~\ref{table:blackBox}. On an average, we observe 30-40\% degradation in the target network's performance. We observe that the generalization is stronger across similar HPE systems, for example- Stacked-Hourglass's perturbation degrades DLCM and Attention-Hourglass to 50\%, but DeepPose and Chained-Prediction to only 75\%. Overall, our study indicates that even under the most restrictive setting, HPE systems can be easily rendered useless.


\subsection{HPE vs. Classification Systems}\label{ssec:exp_hpe_vs_other}
We first compare the robustness of HPE systems in general to another task that involves per-pixel reasoning, semantic segmentation (presented in ~\cite{Arnab_2018_CVPR}).
A simple comparison between the relative-PCKh drop in the performance for FGSM-U attack on HPE $\sim 30$ (Fig. \ref{fig:allplots}d) vs relative IoU drop semantic segmentation $\sim 80$ (ref.~\cite{Arnab_2018_CVPR} Fig. 2(a)) reveals that the HPE systems undergo much less degradation. This fact can be further bolstered by the observation presented in Sec.~\ref{ssec:exp_iterative}, where we observed that the number of iterations required for complete degradation of HPE performance is significantly higher as compared to semantic segmentation system. 
While some part of the observed relative robustness can be attributed to a more lenient metric, PCKh vs. IoU. We believe that some of it perhaps comes from the successive down-sampling and up-sampling of the Stacked-Hourglass introduces multi-scale processing, which has been previously reported to be effective against adversarial attacks on semantic segmentation \cite{Arnab_2018_CVPR}.

\subsection{Relative Robustness among HPE Systems}\label{ssec:exp_relative}

\begin{figure}[b]
\vspace{-1em}
\centering

    \subfloat[Stacked Hourglass]{
        \includegraphics[width=0.43\linewidth]{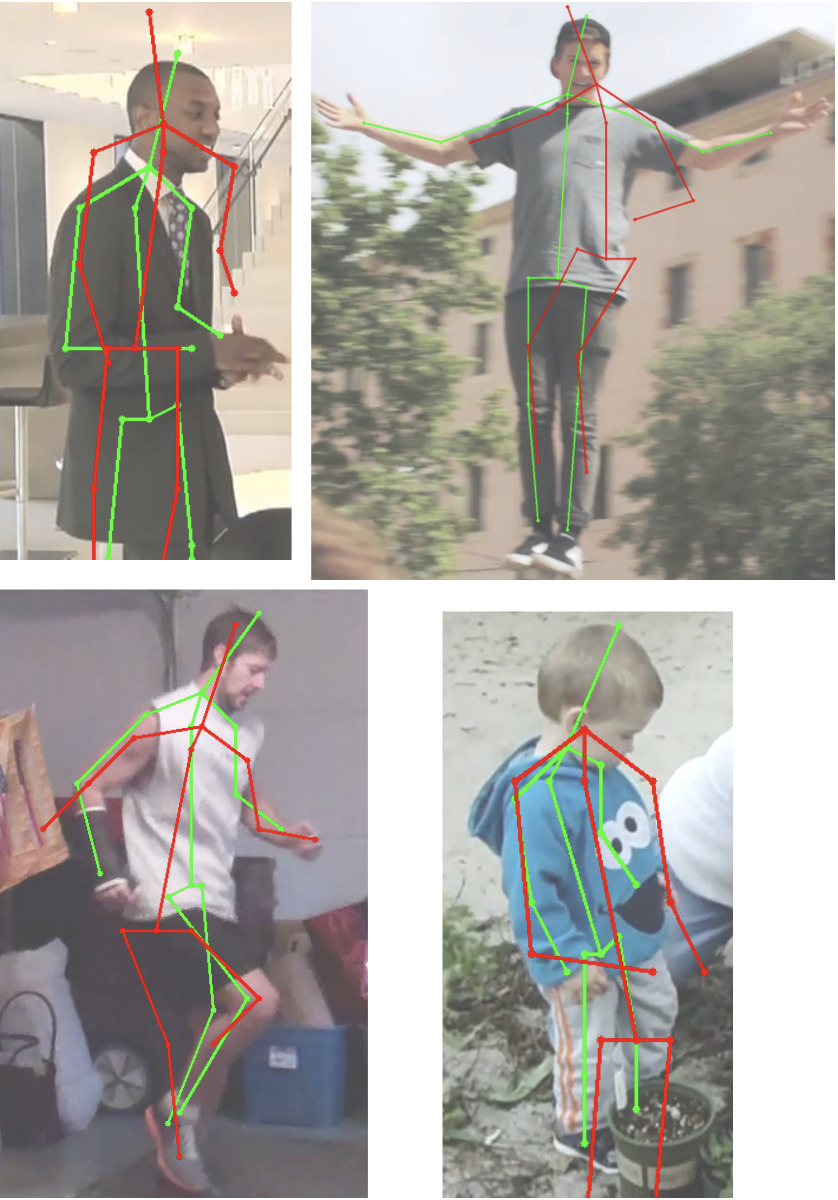}}
    \hfill
    \subfloat[Deep Pose]{
        \includegraphics[width=0.465\linewidth]{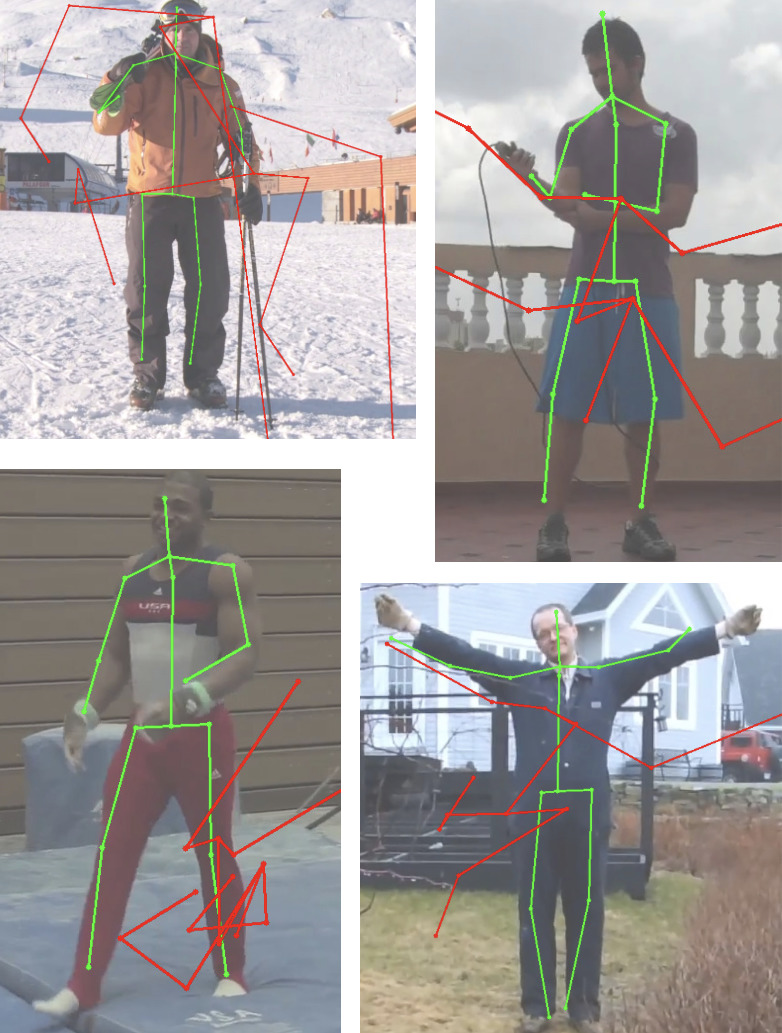}}
    \hfill
    
\caption{Examples of predictions after performing IGSM-U-10 attacks for the Stacked Hourglass and Deep Pose networks. The green and red skeletons represent the predictions of the original and perturbed images, respectively. In all cases, the original correct predictions change to incorrect ones, with the final output of the Deep Pose model having highly unlikely to impossible poses.}
\label{fig:un-targeted_examples}
\end{figure}

A simple inspection of Fig.~\ref{fig:allplots} reveals that the heatmap-based approaches are significantly more robust than the direct-regression based approaches. It could be due to the fact that direct-regression loss function directly translates in to PCKh after thresholding, while heatmap loss produces Gaussian bumps at joint-location, which is not as strongly correlated to PCKh. Moreover, heatmap predictions, unlike regressed values, are implicitly bounded to be valid image coordinates. A few visual examples of the predictions made by the Stacked-Hourglass and DeepPose networks are shown in Fig~\ref{fig:un-targeted_examples} (please refer Fig. \ref{fig:untargeted_vis} in Appendix for visualizations for all models) to further bolster this observation. Somewhat unexpected, the order of robustness among different HPE systems against the attacks is more or less consistent, which indicates that different attack mechanisms are relatively consistent w.r.t. different HPE systems. 
\par In order to make a fair comparison between heatmap and direct regression systems, we use the same ResNet backbone and use a simple regression loss in one case (DeepPose), and de-conv layers followed by heatmap regression in the other case. For the resnet-deconv experiments we further consider two design choices -- with and without ImageNet pre-training. We name them as \textbf{ResDec-Pre} and \textbf{ResDec-NoPre}. As seen in Fig \ref{fig:allplots} (b), relative performance for un-targeted attacks is noticeably higher for heatmap loss (relative-PCKh $=17.3$ for ResDec-No-Pre vs. relative-PCKh $=6.8$ for DeepPose at $\epsilon=8$). This gap in performance is even more evident in FGSM-U attack (Fig. \ref{fig:allplots}d). Also ResDec-Pre performs better than ResDec-No-Pre, validating the findings of ~\cite{pretrain} -- using ImageNet pretraining improves robustness. Strikingly, ResDec-Pre is almost as robust as the most robust network - DLCM. Our findings on non-robustness in direct regression system also advocate a requirement to move away from the popular regression-based 3D-HPE frameworks \cite{Zhou_2017_ICCV, martinez2017, Dabral_2018_ECCV, Hossain_2018_ECCV} (see Sec. \ref{sec:3dhpe} for details on 3D-HPE experiments).
\par Somewhat intuitive, Chained-Prediction heatmap based HPE system turns out to be the least robust against adversarial attack due to the conditional nature of the joint prediction. We observe that DLCM (relative-PCKh 21.6 after IGSM-U-10 attack at $\epsilon=8$) is more robust than 2/8-SHG (15.6 and 18.5 after IGSM-U-10 attack at $\epsilon=8$) against all attacks, perhaps due to DLCM's imposition of human skeleton topology. This encourages further exploration of structure-aware models to counter adversarial attacks. We find that the order of robustness is similar for targeted attacks, with DLCM being the most robust and Chained Predictions and DeepPose being among the most susceptible. However, in this case attacking all hourglasses of the 2-SHG leads to the most potent targeted attack.

\subsubsection{Stacked Hourglass Study} \label{CompSHG}\label{sssec:exp_shg}
Since most HPE systems build on the Stacked-Hourglass backbone ~\cite{NewellStackedEstimation}, we carry out a thorough analysis of adversarial attack on SHG architecture with different network hyper-parameters such as depth (number of stacks) and the position of attack. First, we find that increasing the number of hourglasses from 2 to 8 increases the robustness of the model; a fact clearly visible from Fig. \ref{fig:allplots}(a), \ref{fig:allplots}(d) and \ref{fig:allplots}(f). Next, we study the effect of simultaneous perturbation of outputs of all the stacks of SHG, indicated by suffix \texttt{\(ALL\)}, and observe that the attacks become more effective, again evident from Fig.~\ref{fig:allplots}(a), \ref{fig:allplots}(d) and \ref{fig:allplots}(f). Specifically, 2-SHG-ALL and 8-SHG-ALL attacks increased the target PCKh from 66.3 to 80.5 and from 60.5 to 73.0, respectively. This is expected because downstream stacks are supposed to improve upon the predictions of the upstream ones, therefore, upstream incorrect prediction will cascade into errors in the final output. Furthermore, intermediate supervision will ensure stronger gradient flow down to the input image, especially since the stacks are not connected via residual connections. 
Interestingly, 2-SHG-ALL IGSM-T-20 attack brings down its performance even below Chained-Prediction and DeepPose, the two lowest performing architectures in terms of robustness to adversarial attacks! 

\begin{figure}
\vspace{-1em}
\centering

    \subfloat[Heatmap centred at actual joint location]{
        \includegraphics[width=0.22\linewidth]{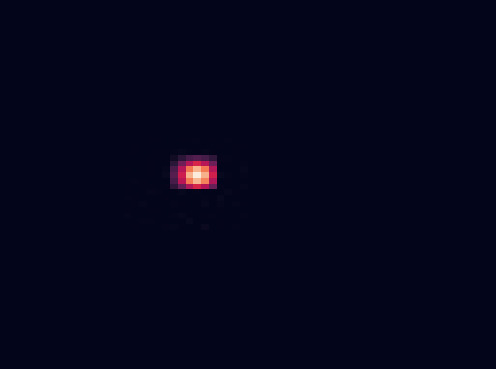}}
    \hfill
    \subfloat[Output for un-targeted Iterative Attack]{
        \includegraphics[width=0.22\linewidth]{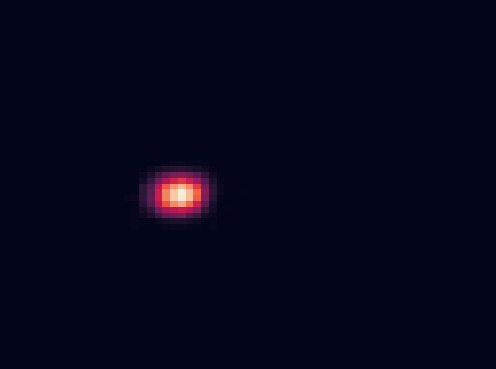}}
    \hfill
    \subfloat[Heatmap centred at target joint location]{
        \includegraphics[width=0.22\linewidth]{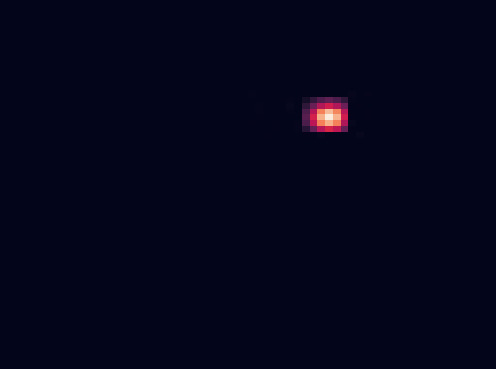}}
    \hfill
    \subfloat[Output for Targeted Iterative Attack]{
        \includegraphics[width=0.22\linewidth]{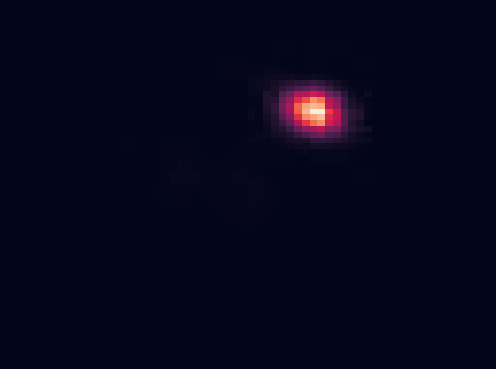}}
    \hfill
    
\caption{Visualization of the various heatmaps produced by the 8-Stack-Hourglass for a particular joint.}
\label{fig:heatmaps}
\end{figure}

\subsubsection {Quality of Heatmaps} \label{sssec:qualityhmp}
Since heatmap-based methods turn out to the most robust against the adversarial attacks, we try to establish whether the characteristics of the Gaussian bumps changes due to the inclusion of adversarial perturbation and potentially serve to detect the attack. Therefore, we follow the approach presented in \cite{Xiao2018CharacterizingSegmentation} and measure the quality of the heatmaps predicted by the HPE systems on adversarial and non-adversarial inputs. We employ KL-divergence as the measure of difference between the Gaussian bumps centered at the location predicted by the system under attack and without the attack i.e. on the original unperturbed images. In all cases we treat the heatmaps produced by the network as the un-normalized log probabilities and compute the KL-divergence of the model's outputs with respect to the ideal Gaussian bumps similar to those used as labels when training. We find that the KL divergence (averaged across all the images in the validation set for the 8-Stacked-Hourglass) is 0.000902 for the un-targeted attacks, 0.001024 for targeted attacks and 0.000640 for the original predictions. This study indicates that even under the adversarial attack the characteristics of the predicted Gaussian bumps have not changed as compared to the original predictions. The fact that the targeted attack yields a higher KL-divergence than the un-targeted is because in many cases the output of the targeted attack contains 2 Gaussian bumps - one centered at the correct joint location and one centered at the target joint location. This can also be visually verified from Fig~\ref{fig:heatmaps}, the heatmaps generated by the model on targeted and un-targeted adversarial samples are similar to the ideal heatmaps. Therefore, such an strategy cannot be reliably employed to detect the presence of an adversarial attack.

\begin{table*}[t]
\centering
 \begin{tabular}{c c c c c c c c c}
\hline
Model and Attack &Ankle &Knee &Hip &Neck &Head &Shoulder &Elbow &Wrist\\[0.5ex]
\hline \hline
& \multicolumn{5}{c}{\textbf{relative-PCKh}} \\
\hline
DeepPose-UI& \textbf{0.63}&1.24&4.43&\underline{17.52}&13.11&4.39&2.35&2.2\\
2-SHG-UI&3.82&4.62&\textbf{2.82}&\underline{41.89}&23.39&24.79&14.48&13.4\\
8-SHG-UI&8.79&10.9&\textbf{3.04}&\underline{45.54}&34.61&29.67&20.65&20.54\\
Chained-UI&2.79&\textbf{2.07}&3.53&\underline{22.7}&15.73&11.87&4.05&3.77\\
Attn-HG-UI&6.52&7.61&\textbf{3.05}&\underline{39.31}&25.01&21.35&17.54&16.96\\
DLCM-UI &6.28&6.79&\textbf{2.12}&\underline{45.69}&29.72&28.04&17.75&16.4\\
\textit{Average} &4.81 &5.54 &\textbf{3.12} &\underline{35.44} &23.60 &20.02 &12.80 &12.22 \\
\hline \hline
& \multicolumn{5}{c}{\textbf{Target PCKh}} \\
\hline
DeepPose-TI&\underline{59.22}&73.04&\textbf{84.79}&81.64&73.0&82.4&77.93&69.33\\
2-SHG-TI &62.61&69.65&\textbf{86.0}&70.05&\underline{49.79}&72.73&64.68&47.59\\
8-SHG-TI &48.24&54.02&\textbf{83.86}&71.94&51.38&70.53&56.33&\underline{43.55}\\
Chained-TI &70.9&77.53&\textbf{84.59}&74.64&\underline{59.29}&75.26&72.28&60.2\\

Attn-HG-TI &47.25&52.06&\textbf{77.97}&60.46&\underline{39.53}&57.22&52.59&48.62\\
DLCM-TI &47.8&54.99&\textbf{74.63}&57.93&\underline{38.88}&55.26&48.58&39.57\\
\textit{Average} &56.00 &63.55 &\textbf{81.97} &69.44 &\underline{51.97} &68.9 &62.07 &\underline{51.47}\\
\hline
\end{tabular}

\caption{relative-PCKh of different body-joints for un-targeted attacks across different networks. \textbf{Boldface} and \underline{underlined} numbers indicate the most and the least vulnerable joints, respectively. Note that hips, knee and ankles are more vulnerable than the rest.}
\label{table:joints}
\end{table*}

\subsection{Body-Joint Vulnerability Towards Attack}
While we have witnessed that the adversarial attack is overall detrimental to HPE systems, we have not yet established which body-joints are more vulnerable than the others. Such a study can help develop special approaches to guard against the most robust body-joints. Therefore, we report per-joint accuracy under different architectures and attack-types for MPII dataset in Table~\ref{table:joints}. For left-right symmetric body-joints (ankle, knee, hip, shoulder, elbow and wrist), we report the left-right average degradation. It's evident that neck and head, with relative-PCKh 35.4 and 23.6, respectively, are the most robust joints while the hips and legs are the most vulnerable across different attacks with relative-PCKh $\leq 5$. It could be due to the fact that the HPE networks are trained on cropped images that have tightly localized head in most of the samples, whereas limbs are spread throughout the images at different locations. Therefore, it is difficult to fool the network in predicting head and neck in some other region. Moreover, we observe that the relative performance of different joints vary dramatically for un-targeted attacks while it it doesn't vary so much for the targeted attacks.

\subsection{Analysis on Multi-Person 2D-HPE Systems (MHPE)} \label{sec:mhpe}
In order to benchmark the robustness for the MHPE problem, we use the standard COCO dataset \cite{coco} and report the MAP values based on the OKS metric. Due to limited computational resources and time-constraints, we report our analysis on the first thousand images from the validation set only. Following the protocols introduced in our experiments for single-person 2D-HPE systems, we switch-off multi-scale inference and left-right flipping to obtain a baseline performance of 63.04 AP and 70.50 AP for our bottom-up \cite{bottom_up_hrnet} and top-down \cite{top_down_hrnet} models, respectively. Similar to the single-person 2D-HPE experiment, we measure the relative-MAP to measure the degradation caused by adversarial attacks.

\begin{figure}[!t]
\centering
\subfloat[FGSM-8 : Bottom-Up Higher HR-Net ]{\includegraphics[width=0.49\linewidth]{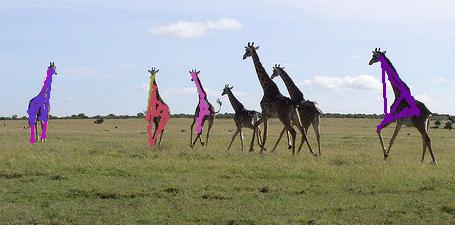}}
\hfill 
\subfloat[FGSM-8 : Top-Down HR-Net model]{\includegraphics[width=0.49\linewidth]{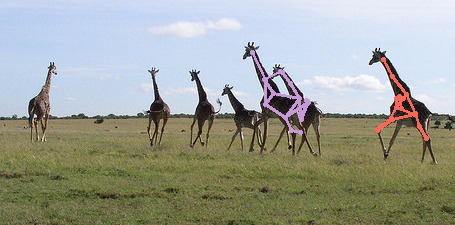}}

\caption{Attacking multi-person 2D-HPE systems -- attack on both top-down and bottom-up networks fool them into predicting humans on Giraffes! Original prediction did not contain humans for either approaches.}
\label{fig:mhpe_attack_img}
\end{figure}

The results of our analysis are summarized in Table \ref{tab:multi_hpe}. From the results, we can conclude that both FGSM and IGSM attacks cause the MAP to fall significantly. In fact IGSM attack can drive the MAP $\sim 0$, which indicates the MHPE systems are more vulnerable than single-person HPE systems. It makes intuitive sense, because now there are two modes of failure -- human detection and key-point detection. In order to visually illustrate this point, Fig.~\ref{fig:mhpe_attack_img} shows images with Giraffes predicting human skeletons by both bottom-up and top-down approaches! The predictions on un-corrupted original image did not output a single human for either of the approaches for this example. Unlike single-person HPE, by definition MHPE doesn't restrict the number of predicted humans, therefore, arbitrary number of human predictions can be generated to reduce the MAP.

\begin{table}[]
    \centering
    \begin{tabular}{c c c c c c c c c c c}
    \hline
    Model and Attack & 0.25 & 0.5 & 1 & 2 & 4 & 8 & 16 & 32 \\
    \hline
    \hline
    & \multicolumn{6}{c}{\textbf{Relative MAP}} \\
    \hline
     Bottom-Up-FGSM & 82.39 & 75.31 & 66.93 & 59.30 & 53.81 & 52.01 & 49.07 & 31.59 \\
     Bottom-Up-IGSM & 48.61 & 21.90 & 7.84 & 2.62 & 1.16 & 0.65 & 0.41 & 0.06 \\ 
     Top-Down-FGSM & 92.34 & 87.66 & 82.13 & 77.87 & 73.19 & 69.50 & 64.40 & 51.06   \\ 
     Top-Down-IGSM & 87.94 & 73.19 & 52.48 & 29.50 & 14.75 & 6.95 & 2.98 & 1.56 \\
    \hline
    \hline
    & \multicolumn{6}{c}{\textbf{Human counts}} \\
    \hline
     Bottom-Up-FGSM & 2.72 & 3.07 & 3.66 & 4.10 & 4.34 & 3.87 & 3.17 & 1.90 \\ 
     Bottom-Up-IGSM & 6.07 & 10.85 & 21.02 & 46.71 & 94.15 & 119.01 & 186.21 & 209.40 \\ 
     Top-Down-FGSM & 7.24 & 7.83 & 8.73 & 9.58 & 10.01 & 10.31 & 9.92 & 8.64 \\ 
     Top-Down-IGSM & 7.40 & 8.39 & 9.73 & 10.89 & 11.95 & 12.29 & 12.14 & 11.19   \\ 
    \hline
    \end{tabular}
    \caption{Relative-MAP computed over the OKS metric and human counts for top-down and bottom-up models under FGSM-U and IGSM-U-10 attacks. Note that for top-down model we only attack the object detection part of the network}
    \label{tab:multi_hpe}
\end{table}

\begin{figure}[!t]
\centering
\subfloat[Original prediction]
{\includegraphics[width=0.32\linewidth]{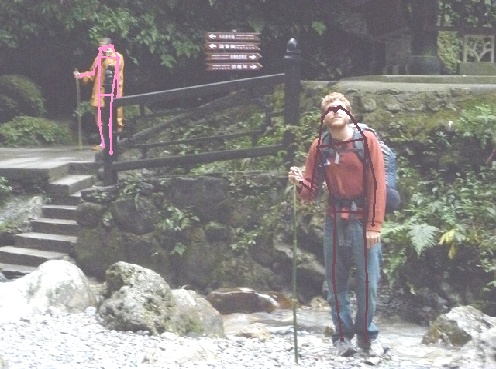}}
\hfill 
\subfloat[Original heatmaps]{\includegraphics[width=0.32\linewidth]{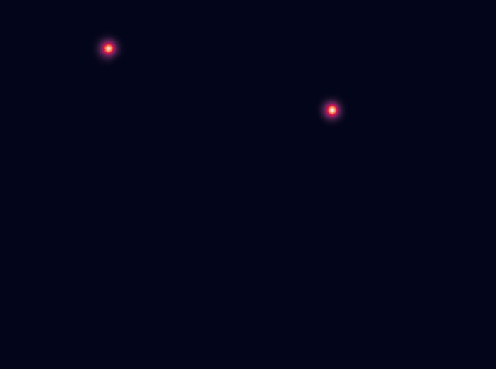}}
\hfill 
\subfloat[Post attack heatmaps ($\epsilon = 8$)]{\includegraphics[width=0.32\linewidth]{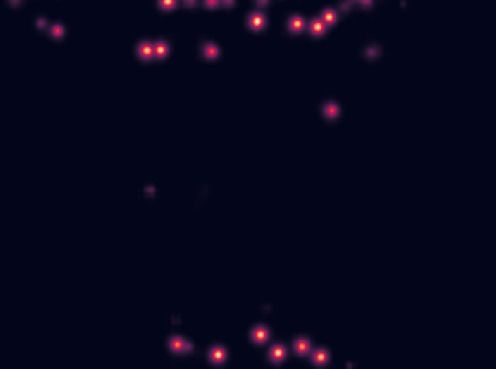}}

\caption{Iterative attacks on bottom up network -- Originally the network correctly identified the two humans in the image but on performing IGSM-U-10 bottom up network tends to produce lot of gaussian bumps in heatmap in Fig. (c) and thereby a lot of humans are predicted. Original heatmap for corresponding joint shown in (b)}
\label{fig:mhpe_heatmaps}
\end{figure}

Since human predictions in the top-down model solely depends on the object-detection module, we can compare the number of humans predicted after the attack as a measure of relative degradation between the two approaches. Our analysis suggests that the bottom-up approaches can be made to predict surprisingly high number of humans under adversarial attack compared to top down methods. The ratio increment in the number of humans is presented in Table~\ref{tab:multi_hpe}. We can see that under IGSM-U-10 attack, the number of humans predicted by bottom-up and top-down approaches increased by $\sim60\times$ vs $2\times$, respectively, for $\epsilon = 8$. Here, we would like to mention that the un-corrupted top-down approach predicted thrice the number of humans (6.57 per image on average) vs. the bottom-up approach (2.04) while the ground-truth number of humans is $\sim2.29$ per image. To visually illustrate this point, we refer to Fig.~\ref{fig:mhpe_heatmaps} for predictions of bottom up method, where -- Fig.~\ref{fig:mhpe_heatmaps}(a) shows the the un-corrupted image with overlaid predictions, Fig.~\ref{fig:mhpe_heatmaps}(b) shows the heatmap for the nose-location on the un-corrupted image, and Fig.~\ref{fig:mhpe_heatmaps}(c) shows the nose-location heatmap for corrupted image for IGSM-U-10 at $\epsilon = 8$. Evidently, the corrupted image gives rise to a heatmap that contains a large number of nose locations. Since the bottom up methods don't restrict the number of persons, it makes them extremely vulnerable to attacks. Interestingly, the Gaussian bumps in the heatmaps still resemble closely to the ideal Gaussian bumps similar to findings in Sec. \ref{sssec:qualityhmp}!

\vspace{-3pt}
\subsection{Analysis of Single-Person 3D-HPE Systems} \label{sec:3dhpe}
Since our selected approach \cite{Zhou_2017_ICCV} for analysis employs a direct-regression on top of the predicted 2D-joint locations to obtain relative depth-map estimation, we simply attack the depth-regressor branch of the network only. This choice also teases apart the relative vulnerability of depth-regressor branch w.r.t. adversarial separately, which is more useful given the aforementioned comprehensive analysis of the 2D-HPE systems already. We perform the IGSM-U-10 attack with $\epsilon = 8$ and report that the MPJPE increased to 360 from the original 60, a six-fold increse in error! For reference, a randomly initialized model yields MPJPE scores between 300-400. Therefore, we can conclude that 3D-HPE systems are also easily fooled by adversarial attacks and that attacking the depth-regression branch only can lead to extremely poor performance. It's likely due to the already discussed and empirically shown strong vulnerability of direct-regressions systems.

\subsection{Human-Perceptibility of Adversarial Perturbation}
In this section, we investigate the extent of visual perceptibility of the adversarial perturbations employed to attack the HPE systems. We perform a user-study, in which the participants were asked to look at a pair of original and corrupted images and indicate whether the two are distinguishable from each other.

We follow protocol of a prior work \cite{FezzaPercept} to conduct this study. We provide 36 people with 31 image pairs with the original image on the left and the corrupted image on the right. The participants were asked to rate the level of difference between the original and the corrupted image. A 3-level grading system was used for rating - 1) imperceptible or similar images, 2) perceptible difference but not annoying, 3) perceptible difference and annoying. We restrict our study to $\epsilon \geq 1$ because for $\epsilon \leq 1$ the difference would be lost due to display system's quantization to integers.

For each of the image pairs, we compute MOS values (mean opinion score), which is the average rating submitted by the users and report in Table~\ref{table:percept}. We can easily witness that increase in $\epsilon$ is strongly correlated with the human-perceptibility of the adversarial perturbation. For $\epsilon \in [1,2,4,8]$, MOS scores is less than 2. For $epsilon \in \{16,32\}$ MOS score is above 2.5. The trend in MOS values can be found in Table \ref{table:percept} which provides, in addition to the MOS values the probability of a randomly chosen image being classified as ``annoying" or ``having non-annoying perceptible differences" by a randomly chosen person, under the assumption that the score assigned is drawn from a Normal distribution, whose mean and variance we estimate. As we have shown IGSM with $\epsilon = 8$ is able to strongly attack all models. Therefore we conclude that even non ``annoying" changes in image can cause network to be fooled. 

\begin{table}
\centering
 \begin{tabular}{|c | c | c |  c|} 
 

 \hline
 Epsilon & MOS & Probab. of being perceptible  & Probab. of being annoying \\ 
 \hline
  1 &  1.14 & 1.03\% & $2.7\times 10^{-5}$\% \\
  2 &  1.15 & 1.12\% & $3.4\times 10^{-5}$\% \\
  4 & 1.44 & 17.34\% & 0.42\%  \\
  8 & 1.96 & 47.93\% & 9.68\%  \\
  16 & 2.54 & 77.87\% & 25.4\%  \\
  32 & 2.86 & 97.84\% & 36.65\%  \\
  \hline
 \end{tabular}
\caption{Results of user-study to determine the perceptibility of attacks. For MOS (Mean Opinion Score), scores of 1, 2 and 3 refer to differences not being perceptible, perceptible differences which are not annoying and perceptible difference which are annoying, respectively. Assuming that the scores provided by a randomly chosen user on a randomly chosen image is a Gaussian for every $\epsilon$}
 \label{table:percept}
 \end{table}

\begin{figure*}[t]
\vspace{-1em}
\centering

    \subfloat[$\epsilon = 1$]{
        \includegraphics[width=0.30\linewidth]{perc_f_1.jpg}}
    \hfill
    \subfloat[$\epsilon = 2$]{
        \includegraphics[width=0.30\linewidth]{perc_f_2.jpg}}
    \hfill
    \subfloat[$\epsilon = 4$]{
        \includegraphics[width=0.30\linewidth]{perc_f_4.jpg}}
    \\
    
\vspace{-0.025\linewidth}
    
    \subfloat[$\epsilon = 8$]{
        \includegraphics[width=0.30\linewidth]{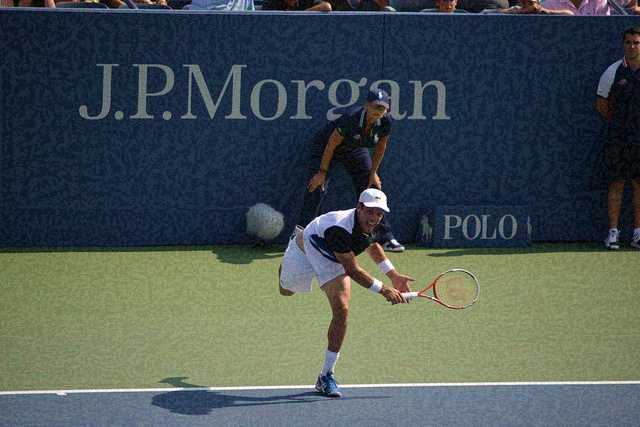}}
    \hfill
    \subfloat[$\epsilon = 16$]{
        \includegraphics[width=0.30\linewidth]{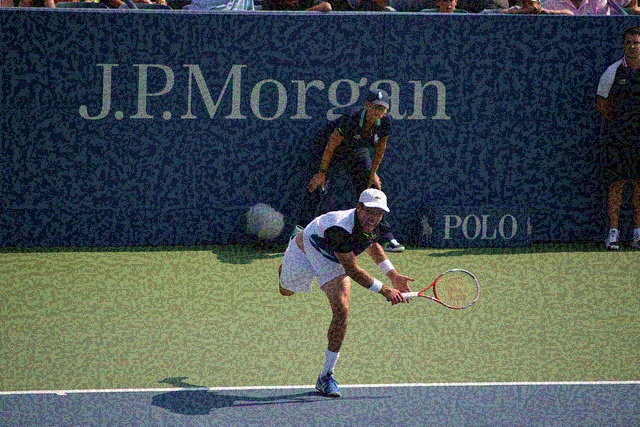}}
    \hfill
    \subfloat[$\epsilon = 32$]{
        \includegraphics[width=0.30\linewidth]{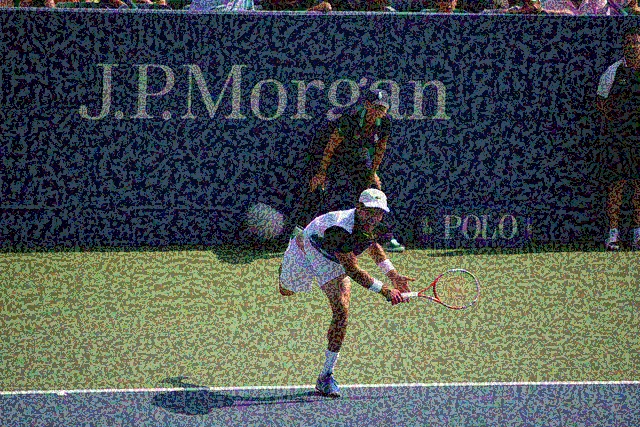}}

\caption{Visualization of adversarial examples of an image at different epsilons. As we can see only for $\epsilon = 16,\ 32$ the perception is impaired. For other values of $\epsilon$ either the effect is not perceptible or it does not cause visual impairment}
\label{fig:UniversalAdversarial}
\end{figure*}

\section{Simple Image Processing for Defense}
In this section we discuss the effect of simple image-processing based defense strategies against adversarial attacks on HPE systems. Since this is a preliminary work on adversarial attacks on human pose, we focus only on computationally cheap methods to mitigate the effect of the above described attacks.

We tried simple geometric and image-processing based defense strategies like flipping and smoothing. As expected, smoothing worked well for both image-specific and image-agnostic attacks, a finding supported by multiple research work in the past \cite{Arnab_2018_CVPR, pixeldef}. Also, we observe that flipping an image-specific perturbations renders it relatively ineffective. Specifically, a non-flipped version of image-specific perturbation degrades the network to a range of 5-10\% whereas, its flipped version can only reduce it to about 70-75\%. This shows that image-specific perturbations are \emph{truly specific} and don't work with flipping. On the other hand, universal perturbations were equally detrimental under flipping too! It can easily be explained on the basis of the fact that universal perturbations are generic while image-dependent perturbation are very specifically aligned. The same is also evident from the visualization of universal perturbations.

\section{Conclusion and Future Work}

\par We performed an exhaustive analysis of various adversarial attacks on single person 2D human pose estimation systems, using MPII ~\cite{andriluka14cvpr} \& COCO ~\cite{coco} and found some interesting trends in how design choices affect robustness. We report that the image-agnostic universal perturbations are as detrimental an attack as image-specific iterative approaches while being computationally much cheaper to obtain. Our visualizations of universal perturbations exhibit a strikingly human-like hallucinated array of body-joints to fool the networks. Further our analyses on the vulnerability of different joints helped identifying the most and least robust body parts under adversarial attack. We finally perform a user study to understand visual impairment caused by these adversarial attacks and found that they are practically feasible.

As a part of future work we would like to explore more human pose specific attacks such as special target poses, person specific attacks such as "t-shirt attacks", "facial feature based attack" etc. Moreover our analysis of 2D MHPE systems and 3D human pose estimation systems opens possible avenues of understanding design choices in those areas.

\begin{acknowledgements}
This work is supported by Mercedes-Benz Research \& Development India (RD/0117-MBRDI00-001). 
\end{acknowledgements}

\bibliographystyle{spbasic}      
\bibliography{egbib}   

\end{document}